\documentclass{article}
\usepackage{arxiv}
\usepackage{framed,multirow}
\usepackage{url}
\usepackage{xcolor}
\definecolor{newcolor}{rgb}{.8,.349,.1}
\usepackage{textcomp}
\usepackage[export]{adjustbox}
\usepackage{amsmath,graphicx}
\usepackage[utf8]{inputenc}
\usepackage{tabularx}
\usepackage{multirow}
\usepackage{interval}
\newcolumntype{L}[1]{>{\raggedright\let\newline\\\arraybackslash\hspace{0pt}}m{#1}}
\newcolumntype{C}[1]{>{\centering\let\newline\\\arraybackslash\hspace{0pt}}m{#1}}
\usepackage[colorlinks=true,citecolor=blue]{hyperref}
\usepackage{siunitx}
\usepackage{placeins}
\usepackage{graphicx}
\usepackage{grffile}
\usepackage{multicol}
\usepackage{float}
\restylefloat{table}
\usepackage{hhline}
\usepackage{cite}
\usepackage[numbers]{natbib}
\usepackage{dsfont}
\usepackage{amssymb}
\newlength{\tempdima}
\newcommand{\rowname}[1]
{\rotatebox{90}{\makebox[\tempdima][c]{#1}}}
\usepackage{pdflscape}
\usepackage{scalerel}
\usepackage{booktabs}

\usepackage{graphicx}
\usepackage{caption}
\usepackage{subcaption}
\usepackage{tikz}
\usepackage{pgfplots}
\usetikzlibrary{spy,calc}
\usepackage{hyperref}

\newif\ifblackandwhitecycle
\gdef\patternnumber{0}

\pgfkeys{/tikz/.cd,
	zoombox paths/.style={
		draw=orange,
		very thick
	},
	black and white/.is choice,
	black and white/.default=static,
	black and white/static/.style={ 
		draw=white,   
		zoombox paths/.append style={
			draw=white,
			postaction={
				draw=black,
				loosely dashed
			}
		}
	},
	black and white/static/.code={
		\gdef\patternnumber{1}
	},
	black and white/cycle/.code={
		\blackandwhitecycletrue
		\gdef\patternnumber{1}
	},
	black and white pattern/.is choice,
	black and white pattern/0/.style={},
	black and white pattern/1/.style={    
		draw=white,
		postaction={
			draw=black,
			dash pattern=on 2pt off 2pt
		}
	},
	black and white pattern/2/.style={    
		draw=white,
		postaction={
			draw=black,
			dash pattern=on 4pt off 4pt
		}
	},
	black and white pattern/3/.style={    
		draw=white,
		postaction={
			draw=black,
			dash pattern=on 4pt off 4pt on 1pt off 4pt
		}
	},
	black and white pattern/4/.style={    
		draw=white,
		postaction={
			draw=black,
			dash pattern=on 4pt off 2pt on 2 pt off 2pt on 2 pt off 2pt
		}
	},
	zoomboxarray inner gap/.initial=5pt,
	zoomboxarray columns/.initial=2,
	zoomboxarray rows/.initial=2,
	subfigurename/.initial={},
	figurename/.initial={zoombox},
	zoomboxarray/.style={
		execute at begin picture={
			\begin{scope}[
				spy using outlines={%
					zoombox paths,
					width=\imagewidth / \pgfkeysvalueof{/tikz/zoomboxarray columns} - (\pgfkeysvalueof{/tikz/zoomboxarray columns} - 1) / \pgfkeysvalueof{/tikz/zoomboxarray columns} * \pgfkeysvalueof{/tikz/zoomboxarray inner gap} -\pgflinewidth,
					height=\imageheight / \pgfkeysvalueof{/tikz/zoomboxarray rows} - (\pgfkeysvalueof{/tikz/zoomboxarray rows} - 1) / \pgfkeysvalueof{/tikz/zoomboxarray rows} * \pgfkeysvalueof{/tikz/zoomboxarray inner gap}-\pgflinewidth,
					magnification=3,
					every spy on node/.style={
						zoombox paths
					},
					every spy in node/.style={
						zoombox paths
					}
				}
				]
			},
			execute at end picture={
			\end{scope}
			\node at (image.north) [anchor=north,inner sep=0pt] {\subcaptionbox*{\label{\pgfkeysvalueof{/tikz/figurename}-image}}{\phantomimage}};
			\node at (zoomboxes container.north) [anchor=north,inner sep=0pt] {\subcaptionbox*{\label{\pgfkeysvalueof{/tikz/figurename}-zoom}}{\phantomimage}};
			\gdef\patternnumber{0}
		},
		spymargin/.initial=0.5em,
		zoomboxes xshift/.initial=1,
		zoomboxes right/.code=\pgfkeys{/tikz/zoomboxes xshift=1},
		zoomboxes left/.code=\pgfkeys{/tikz/zoomboxes xshift=-1},
		zoomboxes yshift/.initial=0,
		zoomboxes above/.code={
			\pgfkeys{/tikz/zoomboxes yshift=1},
			\pgfkeys{/tikz/zoomboxes xshift=0}
		},
		zoomboxes below/.code={
			\pgfkeys{/tikz/zoomboxes yshift=-1},
			\pgfkeys{/tikz/zoomboxes xshift=0}
		},
		caption margin/.initial=4ex,
	},
	adjust caption spacing/.code={},
	image container/.style={
		inner sep=0pt,
		at=(image.north),
		anchor=north,
		adjust caption spacing
	},
	zoomboxes container/.style={
		inner sep=0pt,
		at=(image.north),
		anchor=north,
		name=zoomboxes container,
		xshift=\pgfkeysvalueof{/tikz/zoomboxes xshift}*(\imagewidth+\pgfkeysvalueof{/tikz/spymargin}),
		yshift=\pgfkeysvalueof{/tikz/zoomboxes yshift}*(\imageheight+\pgfkeysvalueof{/tikz/spymargin}+\pgfkeysvalueof{/tikz/caption margin}),
		adjust caption spacing
	},
	calculate dimensions/.code={
		\pgfpointdiff{\pgfpointanchor{image}{south west} }{\pgfpointanchor{image}{north east} }
		\pgfgetlastxy{\imagewidth}{\imageheight}
		\global\let\imagewidth=\imagewidth
		\global\let\imageheight=\imageheight
		\gdef\columncount{1}
		\gdef\rowcount{1}
		
	},
	image node/.style={
		inner sep=0pt,
		name=image,
		anchor=south west,
		append after command={
			[calculate dimensions]
			node [image container,subfigurename=\pgfkeysvalueof{/tikz/figurename}-image] {\phantomimage}
			node [zoomboxes container,subfigurename=\pgfkeysvalueof{/tikz/figurename}-zoom] {\phantomimage}
		}
	},
	color code/.style={
		zoombox paths/.append style={draw=#1}
	},
	connect zoomboxes/.style={
		spy connection path={\draw[draw=none,zoombox paths] (tikzspyonnode) -- (tikzspyinnode);}
	},
	help grid code/.code={
		\begin{scope}[
			x={(image.south east)},
			y={(image.north west)},
			font=\footnotesize,
			help lines,
			overlay
			]
			\foreach \x in {0,1,...,9} { 
				\draw(\x/10,0) -- (\x/10,1);
				\node [anchor=north] at (\x/10,0) {0.\x};
			}
			\foreach \y in {0,1,...,9} {
				\draw(0,\y/10) -- (1,\y/10);                        \node [anchor=east] at (0,\y/10) {0.\y};
			}
		\end{scope}    
	},
	help grid/.style={
		append after command={
			[help grid code]
		}
	},
}

\newcommand\phantomimage{%
	\phantom{%
		\rule{\imagewidth}{\imageheight}%
	}%
}
\newcommand\zoombox[2][]{
	\begin{scope}[zoombox paths]
		\pgfmathsetmacro\xpos{
			(\columncount-1)*(\imagewidth / \pgfkeysvalueof{/tikz/zoomboxarray columns} + \pgfkeysvalueof{/tikz/zoomboxarray inner gap} / \pgfkeysvalueof{/tikz/zoomboxarray columns} ) + \pgflinewidth
		}
		\pgfmathsetmacro\ypos{
			(\rowcount-1)*( \imageheight / \pgfkeysvalueof{/tikz/zoomboxarray rows} + \pgfkeysvalueof{/tikz/zoomboxarray inner gap} / \pgfkeysvalueof{/tikz/zoomboxarray rows} ) + 0.5*\pgflinewidth
		}
		\edef\dospy{\noexpand\spy [
			#1,
			zoombox paths/.append style={
				black and white pattern=\patternnumber
			},
			every spy on node/.append style={#1},
			x=\imagewidth,
			y=\imageheight
			] on (#2) in node [anchor=north west] at ($(zoomboxes container.north west)+(\xpos pt,-\ypos pt)$);}
		\dospy
		\pgfmathtruncatemacro\pgfmathresult{ifthenelse(\columncount==\pgfkeysvalueof{/tikz/zoomboxarray columns},\rowcount+1,\rowcount)}
		\global\let\rowcount=\pgfmathresult
		\pgfmathtruncatemacro\pgfmathresult{ifthenelse(\columncount==\pgfkeysvalueof{/tikz/zoomboxarray columns},1,\columncount+1)}
		\global\let\columncount=\pgfmathresult
		\ifblackandwhitecycle
		\pgfmathtruncatemacro{\newpatternnumber}{\patternnumber+1}
		\global\edef\patternnumber{\newpatternnumber}
		\fi
	\end{scope}
}

\title{HistoStarGAN: A Unified Approach to Stain Normalisation, Stain Transfer and Stain Invariant Segmentation in Renal Histopathology}

\author{
 Jelica Vasiljevi\'{c} \\
  ICube, University of Strasbourg, CNRS (UMR 7357), France \\
  Faculty of Science, University of Kragujevac, Kragujevac, Serbia\\
  \texttt{jvasiljevic@unistra.fr} \\
  \And
 Friedrich Feuerhake \\
  Institute of Pathology, Hannover Medical School, Germany\\
  University Clinic, Freiburg, Germany\\
  \texttt{Feuerhake.Friedrich@mh-hannover.de}
  \And
 C\'{e}dric Wemmert \\
  ICube, University of Strasbourg, CNRS (UMR 7357), France\\
  \texttt{wemmert@unistra.fr}
  \And        
 Thomas Lampert \\
  ICube, University of Strasbourg, CNRS (UMR 7357), France\\
  \texttt{lampert@unistra.fr}
}

\begin{document}
\maketitle
\begin{abstract}

Virtual stain transfer is a promising area of research in Computational Pathology, which has a great potential to alleviate important limitations when applying deep-learning-based solutions such as lack of annotations and sensitivity to a domain shift.  However, in the literature, the majority of virtual staining approaches are trained for a specific staining or stain combination, and their extension to unseen stainings requires the acquisition of additional data and training.  In this paper, we propose HistoStarGAN, a unified framework  that performs stain transfer between multiple stainings, stain normalisation and stain invariant segmentation, all in one inference  of the model. We demonstrate the  generalisation abilities of the proposed solution to perform diverse stain transfer and accurate stain invariant segmentation over numerous unseen stainings, which is the first such demonstration in the field. Moreover, the pre-trained HistoStarGAN model can serve as a synthetic data generator, which paves the way for the use of fully annotated synthetic image data to improve the training of deep learning-based algorithms. To illustrate the capabilities of our approach, as well as the potential risks in the microscopy domain, inspired by applications in natural images, we generated KidneyArtPathology, a fully annotated artificial image dataset for renal pathology.  

\end{abstract}

\section{Introduction}
The deep learning revolution \cite{TheDeepLearningRevolution_2018} opens the door for remarkable applications in the medical domain. Plenty of routine clinical tasks have great potential to be fully automatised, which triggered a staggering amount of research \cite{Piccialli_2021, varoquaux2021failed, liu2019comparison}. In such an environment, Computational Pathology is not an exception \cite{vanderlaakDeepLearningHistopathology2021,Srinidhi2019DeepNN}. However,  currently  many state-of-the-art deep learning methods are data-hungry approaches that require huge collections of annotated data to be trained. However, collecting medical data is strictly regulated, while obtaining high-quality annotations can only be effectively performed only trained experts \cite{grote2018crowdsourcing}. All  this poses important constraints for the development of automated solutions. Moreover, existing data and annotations can  only be reused with limited success,  because the staining process  is prone to high variability \cite{hitology_book}, representing a source of a domain shift  that negatively affects pre-trained models \cite{Tellez2019QuantifyingTE, gadermayr2018which,udagan}.

Generative Adversarial Networks (GANs) \cite{goodfellow2014generative}, being able to generate samples from complex data distributions \cite{karras2020analyzing}, have great potential to overcome some  of the  limitations in applying deep learning-based solutions in Computational Pathology. A particularly promising application area is stain transfer, which enables virtual re-staining of a histopathological image, i.e.\ changing its appearance to look as though it has been stained with another staining or staining variation. Many attempts in the field employ image-to-image translation-based approaches for stain  normalisation \cite{Shaban2019StainganSS,keContrastiveLearningBased2021, DeBel2021},  i.e.\ reducing the  domain shift caused by intra-stain variation, which is a variation in the appearance of the same staining (e.g.\ due to different laboratories procedures). Other approaches try  to reduce inter-stain variation, which is a variation in the appearance of different stainings \cite{Liu2021UnpairedST, gadermayr2018which,levyPreliminaryEvaluationUtility2020,udagan}.  Only a few attempts try to address the generalisation problem of deep-learning-based solutions by proposing a model that is robust to intra-stain variation \cite{wagnerStructurePreservingMultidomainStain2021,xueSelectiveSyntheticAugmentation2021,levine2020synthesis} or inter-stain-variation \cite{udagan}.

However, considering numerous factors of variations which affect the appearance of one staining, in addition to the differences in tissue structures visible under different stainings, the task of virtual stain transfer is naturally non-deterministic and ill-posted. Although state-of-the-art virtual staining strategies, mainly based on GANs, result in visually plausible translations, assessing the results' quality is hard. Obtaining pixel-wise annotations across different staining modalities is limited by the requirement of either processing consecutive tissue sections separately, or re-staining identical sections several times. The latter has the disadvantage of tissue destruction, or processing artefacts. Using consecutive sections that are only stained once  minimises this problem, but the sequentially produced sections are never identical, and their co-registration can be limited by various factors \cite{MERVEILLE2021106157}. Thus, a virtual staining model resulting in multiple translations can support the digital pathology domain to overcome the shortage of high-quality annotations and improve the training of networks. Therefore, it would be reasonable to expect from a virtual staining model to result in diverse translations. However, the approaches which enable such diverse translations are not very common in the field.


In this paper, we propose HistoStarGAN,  a model that can perform diverse translations for stain normalisation and stain transfer tasks, in addition to stain invariant segmentation. The main contributions of this article are:
\begin{itemize}
\item  HistoStarGAN, the first model able to simultaneously perform diverse stain transfer, stain normalisation and stain invariant segmentation. 
\item The proposed model generalises stain transfer across many stainings, including unseen stainings.
\item The model learns using limited annotations from one staining and generalises to a wide range of unseen stainings without requiring additional annotations.
\item KindeyArtPathology -  the first artificially created fully-annotated histological dataset, illustrating the potential and the risks of generating synthetic training data targeting staining methods relevant in renal pathology.  
\item Pre-trained HistoStarGAN, which can be used for numerous offline applications, e.g.\ augmentation of small-size private datasets.
\end{itemize}

The rest of the paper is organised as follows. The HistoStarGAN model is explained in more detail in Section \ref{sec:chapter_histostargan:model_description}; visual and quantitative results are given in Section \ref{sec:chapter_histostargan:subsec_results}. The  HistoStarGAN architecture and training parameters are analysed  in an  ablation study presented in Section \ref{sec:chapter_histostargan:ablation_study}. KindeyArtPathology dataset is explained in more detail Section \ref{sec:chapter_histostargan:sec_kidneyArtPathology_dataset}. The KidneyArtPathology dataset and pre-trained models can be available upon request. Demonstrative examples are provided online\footnote{\href{https://main.d33ezaxrmu3m4a.amplifyapp.com/}{ https://main.d33ezaxrmu3m4a.amplifyapp.com/}}. The limitations and opportunities of the proposed solution are presented in Section \ref{sec:limitations_and_opportunities}; conclusions of the paper are presented in Section \ref{sec:conclusion}.

\section{Related work}

Virtual staining approaches are greatly inspired by style transfer literature \cite{joseGenerativeAdversarialNetworks2021} where GAN-based approaches achieve  state-of-the-art results in many application areas. Therefore, some authors try to bypass the physical staining process, which introduces most of the variation \cite{LiDan2020,ranaUseDeepLearning2020} while others try to normalise slide appearance after staining \cite{Shaban2019StainganSS}. Nevertheless, the methods are usually based on image-to-image translations approaches, where one domain (stain) is translated to look like a reference staining. Some approaches   \cite{salehiPixbasedStaintoStainTranslation2020,choNeuralStainStyleTransfer2017,congSemisupervisedAdversarialLearning2021}  simulate paired datasets by discarding colour information (e.g.\ by using a greyscale version of the image or extracting a haematoxylin channel from the image) and learning paired image-to-image translations models, such as pix2pix \cite{pix2pix2017}. However, discarding the majority of colour information can also eliminate relevant diagnostic details \cite{moghadamStainTransferUsing2022, Bentaieb2018AdversarialST}. Thus,  many approaches  build upon the idea of unpaired image-to-image translation methods,  where CycleGAN-based \cite{CycleGAN2017} approaches are dominant \cite{Shaban2019StainganSS, Kang2020StainNetAF,Shrivastava2021, DeBel2021,loCycleconsistentGANbasedStain2021, Liu2021UnpairedST}. However,  CycleGAN reduces virtual staining to a deterministic mapping, producing a single and fixed output for a given input. Although training repetitions of the same model, or different epochs during training, can result in different translations \cite{vasiljevicCYCLEGANVIRTUALSTAIN2021}, the common characteristic of CycleGAN-based methods is a deterministic translation, which limits output diversity.  Moreover, the method is bi-directional, meaning that its application to more than two stainings requires multiple trainings of the model.  Alternative multi-domain approaches, such as those based on StarGAN \cite{StarGAN2018} are shown to be inferior to CycleGAN in specific applications \cite{udagan}. The quality of more complex models \cite{linUnpairedMultiDomainStain} to reduce stain variation in multi-domain translations remains to be explored. 
Alternative approaches try to overcome the stain variation problem by aiming for stain invariant solutions. Existing methods mainly exploit virtual staining to perform training time augmentation \cite{udagan,wagnerStructurePreservingMultidomainStain2021}, or test-time augmentation \cite{scalbert2022test}. Such a solution can greatly benefit from   diverse virtual staining as a single model can perform more extensive augmentation. 
The current state-of-the-art methods for diverse multi-domain style transfer, such as StarGANv2 \cite{StarGAN2018v2} and TUINT \cite{Baek_2021_ICCV},  could be used for diverse virtual staining.   However, contrary to the domain of natural images where style transfer should alter source-specific image characteristics in the output (e.g.\  the size of ears when translating a cat to a dog,  the amount of hair and hairstyle in female to male transfer etc. ), in the medical domain, such extensive alterations are not desirable as they could result in removing/inventing  specific cell population (like cancerous) or structures, for example glomeruli in renal pathology. Thus, the direct application of current state-of-the-art style transfer models is not straightforward. 
 A recent attempt by \citet{scalbert2022test} employs StarGANv2 translations for test-time augmentation. However, the benefits of this approach are demonstrated in H\&E staining only, and it requires annotated samples from several domains to learn the translation. In  the absence of annotated samples, the training of such  models  becomes difficult, as will be demonstrated in this paper and discussed in more detail in Section \ref{sec:ablation_study_segmenatation_branch}.

This paper proposes an extension of the StarGANv2 model, named HistoStarGAN. The presented approach can perform plausible and diverse virtual staining, preserving the structure of interest during translation. Moreover, the model obtains stain invariant segmentation of the selected structure. The proposed solution results in a single model which is, for the first time, able to perform simultaneous stain normalisation, stain transfer and stain invariant segmentation without any additional data manipulation during test time. The HistoStarGAN is the new state-of-the-art method for stain invariant segmentation, outperforming the current by a large margin. The benefits are demonstrated in several stainings, both histochemical and immunohistochemical, for the task of glomeruli segmentation. 

\section{Method}

\subsection{Model Description}
\label{sec:chapter_histostargan:model_description}

\begin{figure*}[t] 
	\centering 
	\includegraphics[width=\textwidth]{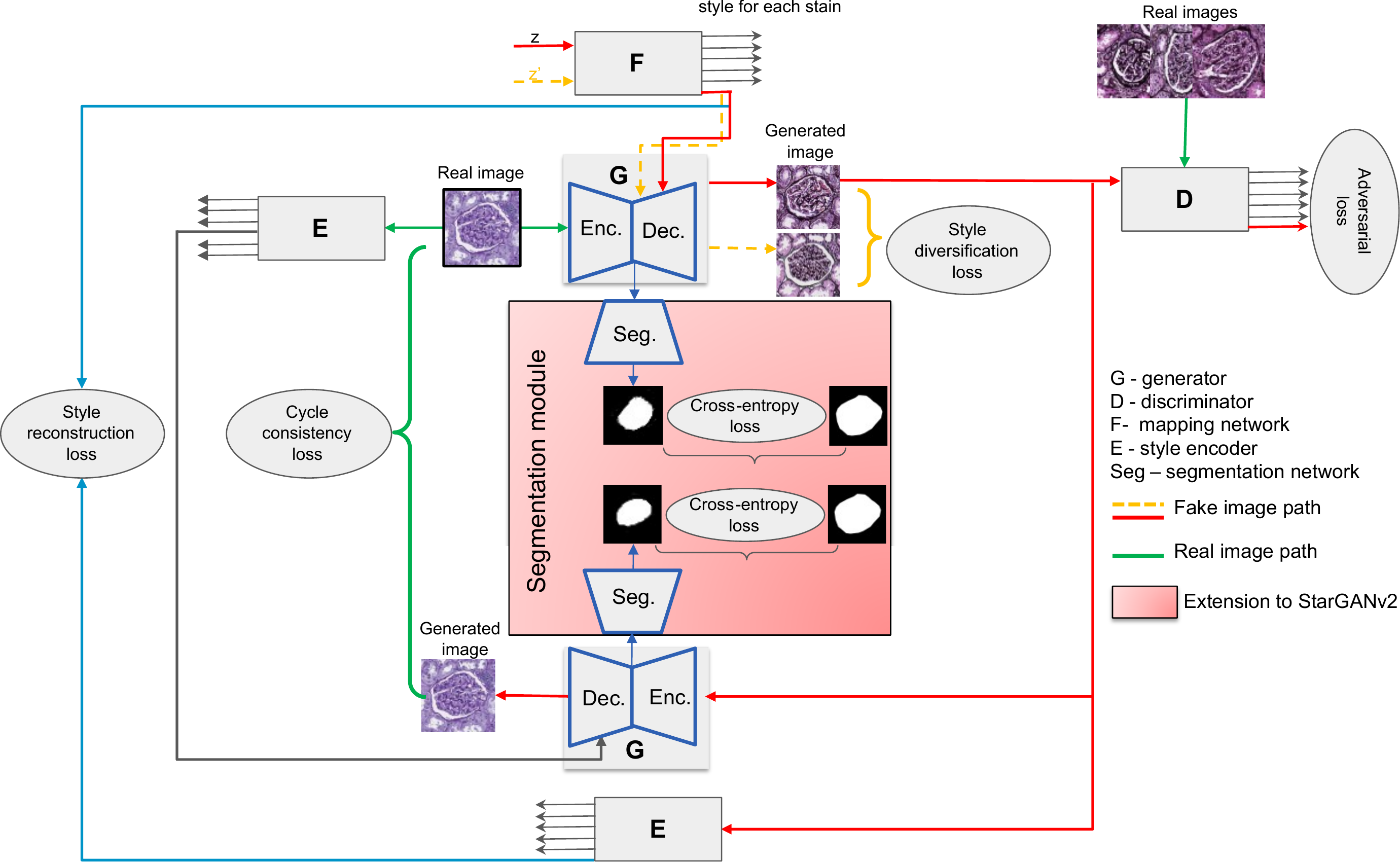}
	\caption{HistoStarGAN -- an end-to-end trainable model for simultaneous stain transfer and stain invariant segmentation. Red block denotes the difference compared to StarGANv2 model \cite{StarGAN2018v2}.} 
	\label{fig:starganv2_diagram}
\end{figure*}

HistoStarGAN architecture is presented in Figure \ref{fig:starganv2_diagram}. The model is composed of five modules: generator ($G$), discriminator ($D$), mapping network ($F$), style encoder ($E$) and segmentation network ($S$). The models $G$, $D$, $F$ and $E$  are elements of  StarGANv2's architecture. 

The mapping network $F$ generates a stain-specific style by transforming the random latent code $z$ into the target stain's style. The style is injected into the generator $G$ during  translation, which enables diverse generations as different latent codes result in different stain-specific styles. In order to ensure that the generator uses the injected style information, the model is constrained with a style reconstruction loss, i.e.\ the style encoder $E$ extracts the style from the generated image, and the difference between that style and the style provided to the generator during translation is minimised (blue arrows in the Figure \ref{fig:starganv2_diagram}).  
In order to explicitly allow style diversification, the model is trained to produce different outputs for different styles in the given target domain by the style diversification loss (orange arrows in the Figure \ref{fig:starganv2_diagram}). Moreover, the model is constrained using a cycle-consistency loss (green brackets in Figure \ref{fig:starganv2_diagram}), e.g.\ the difference between the original and reconstructed images is minimised. Reconstruction is performed using the same generator, but the style information is extracted by the mapping network  by  taking the original image as input.

The generator $G$ is an encoder-decoder network, with an instance normalisation layer in the encoder and an adaptive instance normalisation layer in the  decoder.  In this way,  the encoder removes stain-specific characteristics from the  image while the decoder injects target-stain characteristics during the generation process. Thus, the features extracted by the generator's encoder should  be stain invariant. Under the assumption that a structure of interest is visible in all stainings, i.e.\ the stain invariant solution is feasible, the representation extracted in the bottleneck should be sufficient to perform the considered object-related task. Therefore, a segmentation module is attached to the bottleneck. Trained end-to-end with the other modules, this extension forces the preservation of structures of interest during the translation process. 

More formally, let  $\mathcal{X}$ be the set of histopathological images,  $\mathcal{S}$  the set of available segmentation masks for the structure of interest, and $\mathcal{Y}$ the set of stainings found  in $\mathcal{X}$. Given an image $x \in \mathcal{X}$, its original staining  $y \in \mathcal{Y}$ and corresponding segmentation mask $m_{seg}$, the model is trained using the following objectives.
\paragraph{Adversarial objective:}
The latent code $z \in \mathcal{Z}$, source domain $y\in \mathcal{Y}$ and target domain $\tilde{y}\in \mathcal{Y}$ are randomly sampled. The target style code $\tilde{s}=F_{\tilde{y}}(z)$ is obtained via the mapping network. Thus the generator $G$,  with input image $x$ from domain $y$ and style $\tilde{s}$, generates  and output image $G(x,\tilde{s})$. This image is evaluated by the discriminator's output, which corresponds to a domain $\tilde{y}$, $D_{\tilde{y}}$. These are trained using  the adversarial loss, such that 
\begin{flalign} 
	\mathcal{L}_{\text{adv}} = \mathds{E}_{x,y} [\log{D_y}(x)]+ \mathds{E}_{x,\tilde{y},z} [\log{(1-D_{\tilde{y}}(G(x,\tilde{s})))}].
\end{flalign}

\paragraph{Style reconstruction:}
In order to ensure that generator $G$ uses the provided style code $\tilde{s}$ when producing output $G(x,\tilde{s})$, the following style reconstruction loss is used:
\begin{flalign} 
	\mathcal{L}_{\text{sty}} = \mathds{E}_{x,\tilde{y},z} [\|\tilde{s}-E_{\tilde{y}(G(x,\tilde{s}))}  \|_{1} ].	
\end{flalign} 

\paragraph{Style diversification:}
The generator is forced to produce different outputs for different styles produced by the mapping network $F$. Given different latent codes $z_1$ and $z_2$,  the following diversity loss is maximised:
\begin{flalign} 
	\mathcal{L}_{\text{ds}} = \mathds{E}_{x,\tilde{y},z_1,z_2}[\|G(x,\tilde{s_1}) - G(x,\tilde{s_1})\|_{1} ]. 	
\end{flalign} 

\paragraph{Cycle-consistency:}
The following cycle-consistency loss forces  the generator to reconstruct the original image given the source style code $\hat{s}=E_y(x)$:
\begin{flalign} 
	\mathcal{L}_{\text{cyc}} = \mathds{E}_{x,y,\tilde{y},z}[\|x-G(G(x,\tilde{s}),\hat{s})\|_{1} ]. 	
\end{flalign} 

\paragraph{Segmentation objective:} The cross-entropy loss is used to train the segmentation branch on both real data and their translation to a random stain, such that
\begin{flalign} 
	\mathcal{L}_{\text{seg}} = \mathds{E}_{x,m_{seg}}[m_{seg}\log{Seg(x)} ] +  \mathds{E}_{x,m_{seg},z}[m_{seg}\log{}Seg(G(x,\tilde{s})) ].
\end{flalign}

\paragraph{Full objective:}
The importance of each of these losses is controlled by hyperparameters, and are combined in the following full objective:
\begin{flalign} 
	\min_{G,F,E} \max_{D} \mathcal{L}_{\text{adv}}+\lambda_{\text{sty}}\mathcal{L}_{\text{sty}} -\lambda_{\text{ds}}\mathcal{L}_{\text{ds}} +\lambda_{\text{cyc}}\mathcal{L}_{\text{cyc}} +\lambda_{\text{seg}}\mathcal{L}_{\text{seg}}.
\end{flalign}

\subsection{Training Setup}
\subsubsection{Dataset}
The dataset contains tissue samples collected from a cohort of $10$ patients who underwent tumour nephrectomy due to renal carcinoma. The kidney tissue was selected as distant as possible from the tumours to display largely normal renal glomeruli; some samples included variable degrees of pathological changes such as full or partial replacement of the functional tissue by fibrotic changes (``sclerosis'') reflecting normal age-related changes or the renal consequences of general cardiovascular comorbidity (e.g.\ cardiac arrhythmia, hypertension, arteriosclerosis). The paraffin-embedded samples were cut into $3 \mu m$ thick sections and stained with either Jones' H\&E basement membrane stain (Jones), PAS or Sirius Red, in addition to two immunohistochemistry markers (CD34, CD68), using an automated staining instrument (Ventana Benchmark Ultra). Whole slide images were acquired using an Aperio AT2 scanner at $40\times$ magnification (a resolution of 0.253 $\mu m$ / pixel). All the glomeruli in each WSI were annotated and validated by pathology experts by outlining them using Cytomine \cite{maree2016collaborative}. The dataset was divided into $4$ training, $2$ validation, and $4$ test patients.

Training the HistoStarGAN model's segmentation branch is supervised, and requires the segmentation masks for all images in the dataset. However, in Computational Pathology this assumption is unrealistic.  It is more reasonable to assume that annotations exist for limited examples, e.g.\ as is common in the field, only for one staining  \cite{gadermayr2018which, udagan}. Thus, the PAS staining is considered to be annotated (the source stain), while the  other stainings (Jones H\&E, Sirius Red, CD34, CD68) are considered to be unannotated.

To overcome the lack of annotations in target stainings, the CycleGAN's deterministic nature and limited capacity to perform geometrical changes are used to artificially generate a fully-annotated dataset. The CycleGAN model is trained in an unsupervised manner, using randomly extracted patches from a given pair of stainings. Separate CycleGAN models are trained for each pair of PAS-target stains. When trained, the CycleGAN models are applied to the  source (annotated) dataset in order to generate annotated samples in  the target stainings.

\subsubsection{Training Details}
The CycleGAN models have a  9 ResNet blocks architecture, and they are trained  for $50$ epochs, with patches of size $512 \times 512$ pixels extracted from the training patients in each staining. Upon training, a balanced dataset is formed by translating $500$ glomeruli and $500$  random negative patches from  PAS  to all target stains. When training the HistoStarGAN model,  extensive data augmentation is performed, following the conclusions by \citet{Karras2020ada} that data augmentation is a crucial factor when training GANs with limited data. 

The following augmentations are applied $50\%$ of time with an independent probability of $0.5$ (batches are augmented `on the fly') for each method; elastic deformation  ($\sigma = 10$); affine transformations  -- random rotation in the range $\interval{\ang{0}}{\ang{180}}$, random shift sampled from $\interval{-5}{5}$ pixels, random magnification sampled from $\interval{0.95}{1}$, and horizontal/vertical flip; brightness and contrast enhancements with factors sampled from $\interval{0.0}{0.2}$ and $\interval{0.8}{1.2}$ respectfully;  additive Gaussian noise with $\sigma \in \interval{0}{0.01}$.

The HistoStarGAN model is trained using  the following loss weights: $\lambda_{\text{sty}}=1$, $\lambda_{\text{ds}}=1$, $\lambda_{\text{cyc}}=1$, and $\lambda_{\text{seg}}=5$. To  stabalise training, the weight of style diversification loss, $\lambda_{\text{ds}}$, is linearly decreased to zero over  \num{100000} iterations \cite{StarGAN2018v2}. Also, the weight of the segmentation loss, $\lambda_{\text{seg}}$, is $0$ for the first \num{10000} iterations until the model starts to generate recognisable images from each stain. Although their fidelity is not high enough, at this moment the segmentation model receives enough  meaningful information to start learning.  HistoStarGAN is trained for \num{100000} iterations. The architectures for the generator, discriminator, style encoder, and mapping network are the same as in the original StarGANv2 architecture \cite{StarGAN2018v2} as well as optimisers and learning rates. The segmentation branch's architecture is the same as the generator's decoder, without the adaptive instance normalisation layer. The segmentation branch is trained using the Adam optimiser with a learning rate $10^{-5}$. As for the other networks, exponential moving averages over parameters \cite{Karras2018ProGAN} is applied during training to obtain the final segmentation network, as experimentally, it gives better results than the best model saved based on validation performance.

The HistoStarGAN model's segmentation branch is trained using a balanced dataset, mainly containing artificial histological images produced by both CycleGAN and HistoStarGAN. However, tasks in Computational Pathology are usually concerned with sparse structures, and using an imbalanced dataset to account for the tissue diversity is beneficial for learning \cite{lampert2019strategies,udagan}. Moreover, CycleGAN-based translations can be noisy \cite{vasiljevicCYCLEGANVIRTUALSTAIN2021}, which could affect the stain invariant properties of the segmentation module. Thus, after training HistoStarGAN as a whole, the segmentation module is fine-tuned for one epoch using real unbalanced PAS-stained images while the rest of the model is fixed 
(the choice of the number of fine-tuning epochs is discussed in Section \ref{sec:chapter_histostargan:ablation_study}). This dataset contains all PAS glomeruli ($662$ extracted from the training patients) and seven times more negative patches ($4634$) to account for tissue variability. The Adam optimiser is used with a batch size of $8$ and a learning rate of $0.0001$. The same augmentation as in \citet{lampert2019strategies} is applied during fine-tuning.  
\section{Results}
\label{sec:chapter_histostargan:subsec_results}

This section will demonstrate that  HistoStarGAN  results in a single model able to perform diverse stain transfer, stain normalisation and stain invariant segmentation.  Moreover, having a stain-invariant encoder, the HistoStarGAN model can, for the first time, generalise stain transfer to unseen stainings. 

\subsection{Diverse Multi-Domain Stain Transfer}
\begin{figure}[!htbp]
	\begin{center}
		\settoheight{\tempdima}{\includegraphics[width=0.10\textwidth]{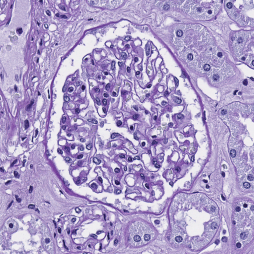}}%
		\begin{tabular}{@{}c@{ }c@{ }c@{ }c@{ }c@{ }c@{ }c@{ }c@{ }c@{ }c@{ }}
			& \multicolumn{5}{c}{Translations}\\
				\cline{2-7} \vspace{-2.5ex}\\
			Real PAS &PAS &Jones H\&E & Sirius Red & CD68 & CD34\\
			\includegraphics[width=0.15\textwidth]{Images_for_HistoStarGANv2/HistoStarGAN_balanced_dataset_w_5/PAS_to_others/02_original.png}&
			\includegraphics[width=0.15\textwidth]{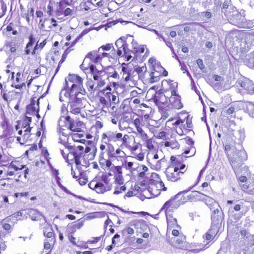} &
			\includegraphics[width=0.15\textwidth]{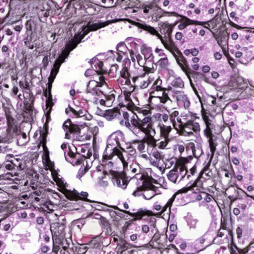} &
			\includegraphics[width=0.15\textwidth]{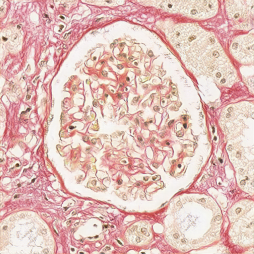} &
			\includegraphics[width=0.15\textwidth]{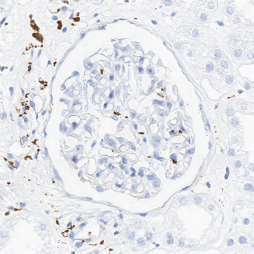} &
			\includegraphics[width=0.15\textwidth]{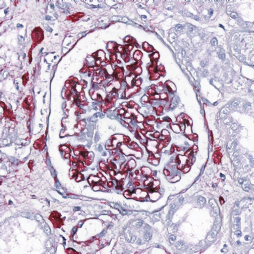} &
			 \\
			 \includegraphics[width=0.15\textwidth]{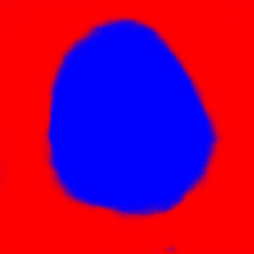}&
			 \includegraphics[width=0.15\textwidth]{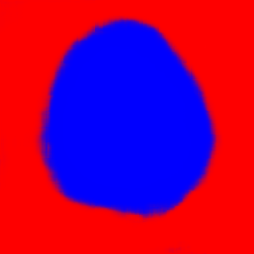} &
			 \includegraphics[width=0.15\textwidth]{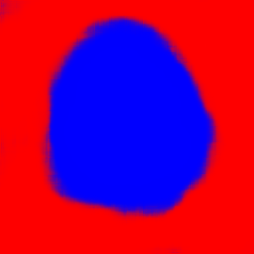} &
			 \includegraphics[width=0.15\textwidth]{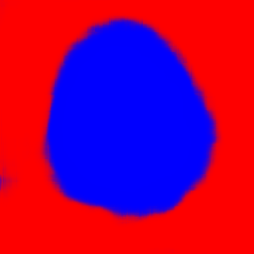} &
			 \includegraphics[width=0.15\textwidth]{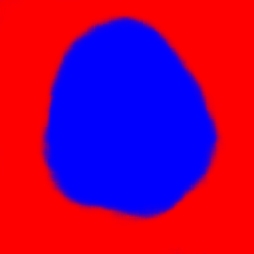} &
			 \includegraphics[width=0.15\textwidth]{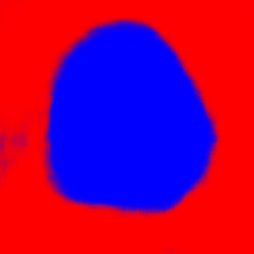} &
			 \\
			&
		\includegraphics[width=0.15\textwidth]{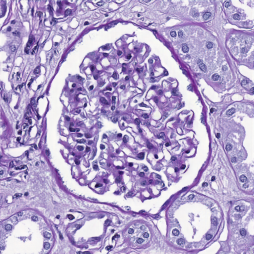} &
		\includegraphics[width=0.15\textwidth]{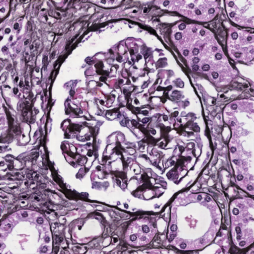} &
		\includegraphics[width=0.15\textwidth]{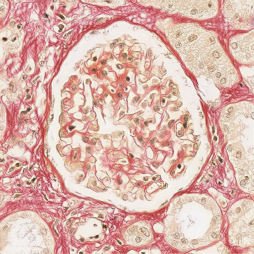} &
		\includegraphics[width=0.15\textwidth]{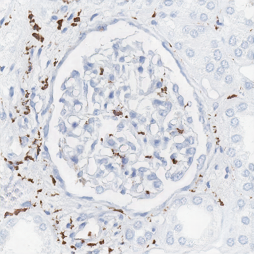} &
		\includegraphics[width=0.15\textwidth]{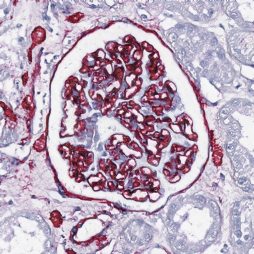} 
		\\
		&
		\includegraphics[width=0.15\textwidth]{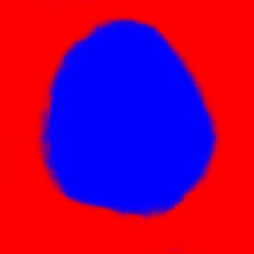} &
		\includegraphics[width=0.15\textwidth]{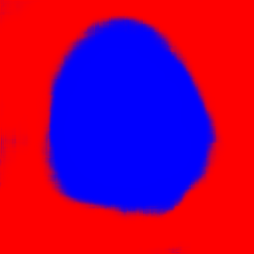} &
		\includegraphics[width=0.15\textwidth]{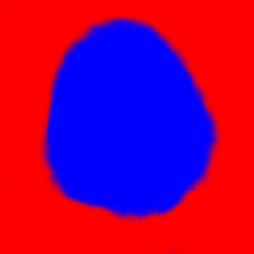} &
		\includegraphics[width=0.15\textwidth]{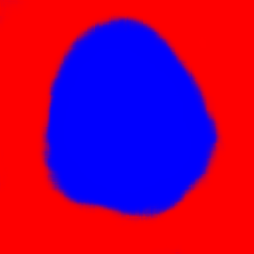} &
		\includegraphics[width=0.15\textwidth]{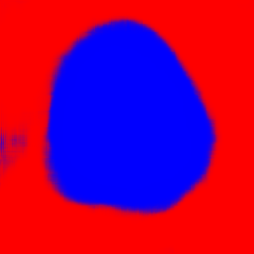} 
		\\
			&
		\includegraphics[width=0.15\textwidth]{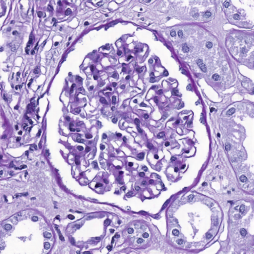} &
		\includegraphics[width=0.15\textwidth]{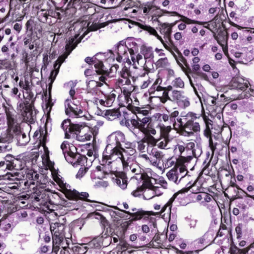} &
		\includegraphics[width=0.15\textwidth]{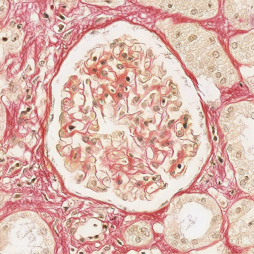} &
		\includegraphics[width=0.15\textwidth]{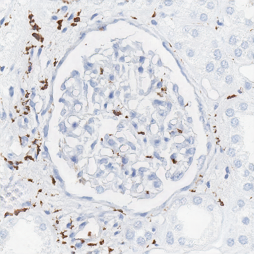} &
		\includegraphics[width=0.15\textwidth]{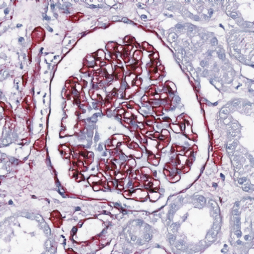} 
		\\
		&
		\includegraphics[width=0.15\textwidth]{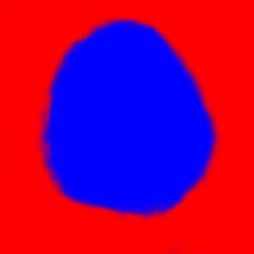} &
		\includegraphics[width=0.15\textwidth]{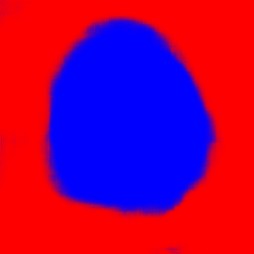} &
		\includegraphics[width=0.15\textwidth]{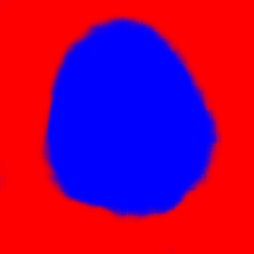} &
		\includegraphics[width=0.15\textwidth]{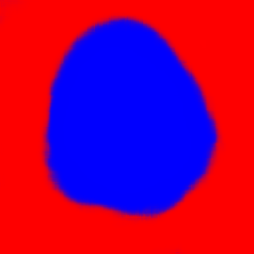} &
		\includegraphics[width=0.15\textwidth]{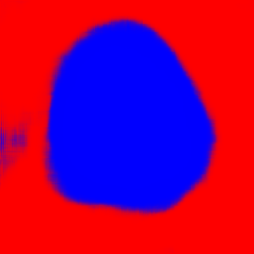} 
		\\
			
		\end{tabular}
		\caption{Diverse HistoStarGAN translations of a PAS glomeruli patch to target stains (including PAS) with corresponding segmentations. Fake images in each row are generated using the same random vector transformed into a stain-specific style by the Mapping network. The differences between translations are in microscopic structures (e.g.\ membrane weight or nucleus appearing), which is barely visible in these figures.  Full resolution images, in which these differences are more visible,  are available  \href{https://github.com/jecaJeca/HistoStarGAN}{online}.
		}
		\label{fig:histostargan_pas_translations}
	\end{center}
\end{figure}

\begin{figure}[!htbp]
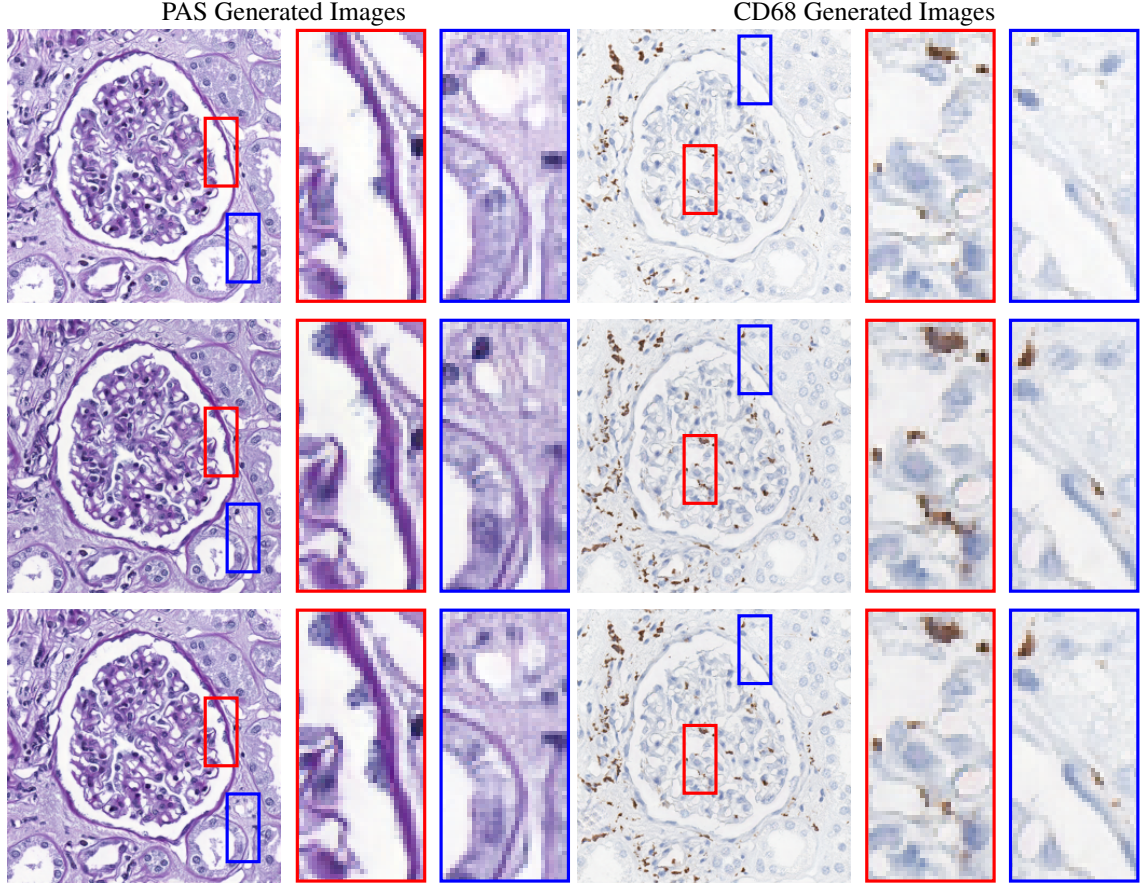

	\begin{center}
		\begin{tabular}{c@{ }c@{ }c@{ }c@{ }}
			\multicolumn{1}{c}{PAS Generated Images} &	\multicolumn{1}{c}{CD68 Generated Images} \\
			\begin{tikzpicture}[
				zoomboxarray,zoomboxarray columns=2,,zoomboxarray rows=1
				]
				\node [image node] { \includegraphics[width=0.22\textwidth]{Images_for_HistoStarGANv2/HistoStarGAN_balanced_dataset_w_5/PAS_to_others/02_02_latent_44.png} };
				\zoombox[color code=red,magnification=4]{0.78,0.55}
				\zoombox[color code=blue,magnification=4]{0.86,0.2}
				
			\end{tikzpicture}
			&
			\begin{tikzpicture}[
				zoomboxarray, zoomboxarray columns=2,,zoomboxarray rows=1,
				]
				\node [image node] { \includegraphics[width=0.22\textwidth]{Images_for_HistoStarGANv2/HistoStarGAN_balanced_dataset_w_5/PAS_to_others/02_16_latent_44.png} };
				\zoombox[color code=red,magnification=4]{0.45,0.45}
				\zoombox[color code=blue,magnification=4]{0.65,0.85}
				
			\end{tikzpicture}

			\vspace{-3ex}\\
			
			\begin{tikzpicture}[
				zoomboxarray, zoomboxarray columns=2,,zoomboxarray rows=1,
				]
				\node [image node] { \includegraphics[width=0.22\textwidth]{Images_for_HistoStarGANv2/HistoStarGAN_balanced_dataset_w_5/PAS_to_others/02_02_latent_42.png} };
				\zoombox[color code=red,magnification=4]{0.78,0.55}
				\zoombox[color code=blue,magnification=4]{0.86,0.2}
				
			\end{tikzpicture} 
			&
			\begin{tikzpicture}[
				zoomboxarray, zoomboxarray columns=2,,zoomboxarray rows=1,
				]
				\node [image node] { \includegraphics[width=0.22\textwidth]{Images_for_HistoStarGANv2/HistoStarGAN_balanced_dataset_w_5/PAS_to_others/02_16_latent_42.png} };
				\zoombox[color code=red,magnification=4]{0.45,0.45}
				\zoombox[color code=blue,magnification=4]{0.65,0.85}
				
			\end{tikzpicture}

			\vspace{-3ex}\\
			
			\begin{tikzpicture}[
				zoomboxarray, zoomboxarray columns=2,,zoomboxarray rows=1,
				]
				\node [image node] { \includegraphics[width=0.22\textwidth]{Images_for_HistoStarGANv2/HistoStarGAN_balanced_dataset_w_5/PAS_to_others/02_02_latent_22.png} };
				\zoombox[color code=red,magnification=4]{0.78,0.55}
				\zoombox[color code=blue,magnification=4]{0.86,0.2}
				
			\end{tikzpicture}
			&
			
			\begin{tikzpicture}[
				zoomboxarray, zoomboxarray columns=2,,zoomboxarray rows=1,
				]
				\node [image node] { \includegraphics[width=0.22\textwidth]{Images_for_HistoStarGANv2/HistoStarGAN_balanced_dataset_w_5/PAS_to_others/02_16_latent_22.png} };
				\zoombox[color code=red,magnification=4]{0.45,0.45}
				\zoombox[color code=blue,magnification=4]{0.65,0.85}
			\end{tikzpicture}
			\vspace{-3ex}
		\end{tabular}
		\caption[ A closer look at the differences between translations]{ A closer look at the differences between translations.}
		\label{fig:histostargan_zppmed_translations}
	\end{center}
\end{figure}

\begin{figure}[!htbp]
	\begin{center}
		\settoheight{\tempdima}{\includegraphics[width=0.15\textwidth]{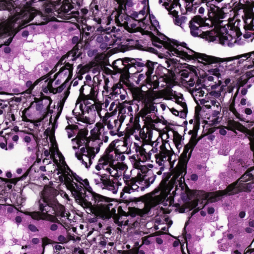}}%
		\begin{tabular}{@{}c@{ }c@{ }c@{ }c@{ }c@{ }c@{ }c@{ }c@{ }c@{ }c@{ }}
			& & \multicolumn{5}{c}{Translationss}\\
				\cline{3-7} \vspace{-2.5ex}\\
			& Real &PAS &Jones H\&E & Sirius Red & CD68 & CD34\\
			\rowname{Jones H\&E} &
			\includegraphics[width=0.15\textwidth]{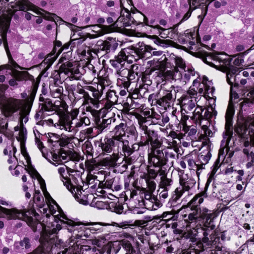}&
			\includegraphics[width=0.15\textwidth]{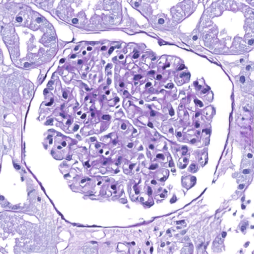}&
			\includegraphics[width=0.15\textwidth]{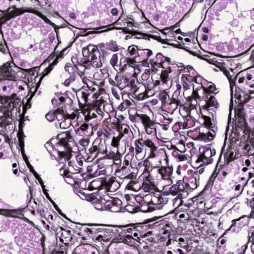} &
			\includegraphics[width=0.15\textwidth]{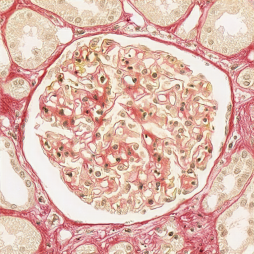} &
			\includegraphics[width=0.15\textwidth]{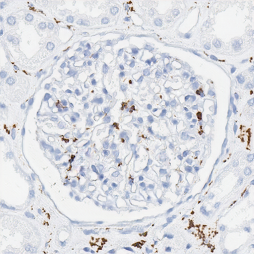} &
			\includegraphics[width=0.15\textwidth]{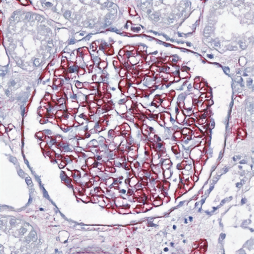}
			\\
			\rowname{Seg} &
			\includegraphics[width=0.15\textwidth]{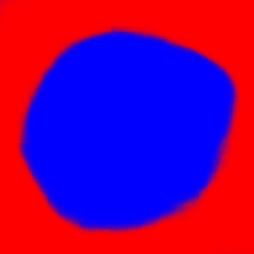}&
			\includegraphics[width=0.15\textwidth]{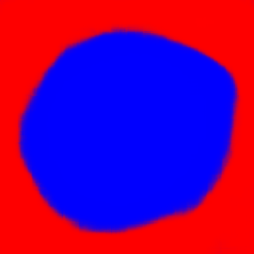}&
			\includegraphics[width=0.15\textwidth]{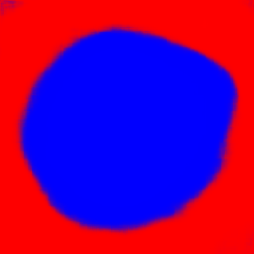} &
			\includegraphics[width=0.15\textwidth]{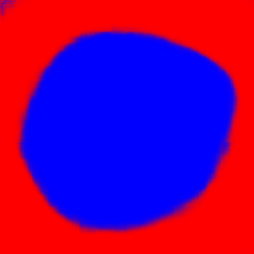} &
			\includegraphics[width=0.15\textwidth]{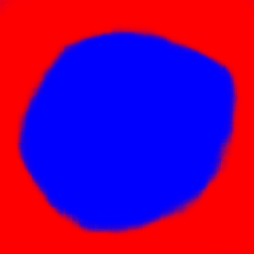} &
			\includegraphics[width=0.15\textwidth]{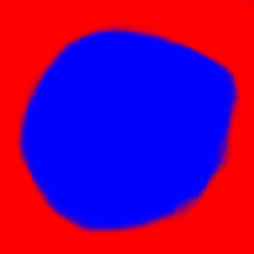}
			\\
			\rowname{CD68} &
			\includegraphics[width=0.15\textwidth]{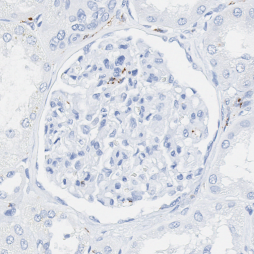}&
			\includegraphics[width=0.15\textwidth]{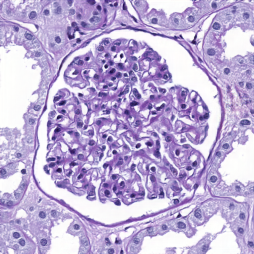}&
			\includegraphics[width=0.15\textwidth]{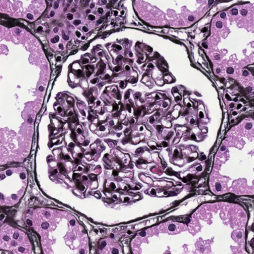} &
			\includegraphics[width=0.15\textwidth]{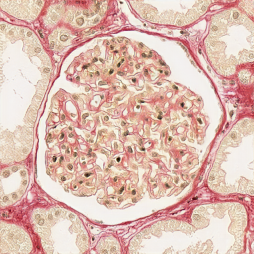} &
			\includegraphics[width=0.15\textwidth]{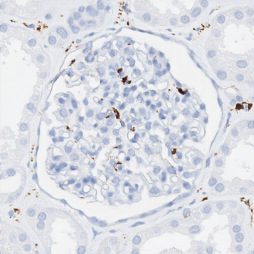} &
			\includegraphics[width=0.15\textwidth]{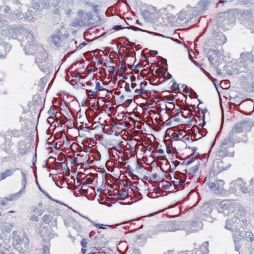}
			\\
				\rowname{Seg.} &
			\includegraphics[width=0.15\textwidth]{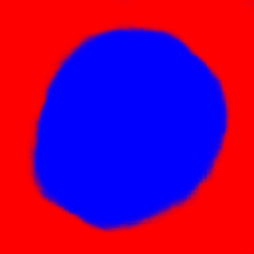}&
			\includegraphics[width=0.15\textwidth]{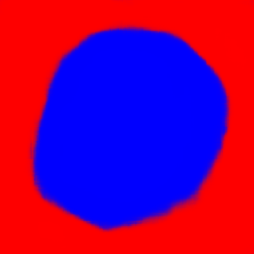}&
			\includegraphics[width=0.15\textwidth]{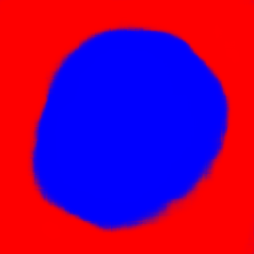} &
			\includegraphics[width=0.15\textwidth]{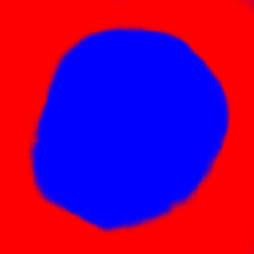} &
			\includegraphics[width=0.15\textwidth]{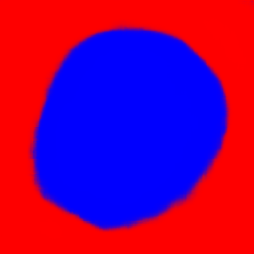} &
			\includegraphics[width=0.15\textwidth]{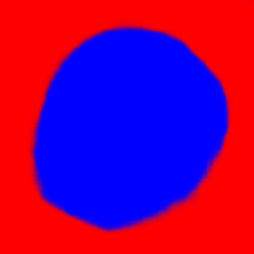}
			\\
			\rowname{Sirius Red} &
			\includegraphics[width=0.15\textwidth]{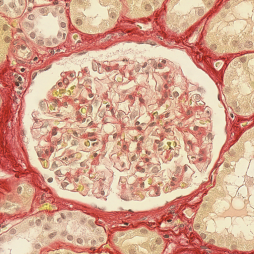}&
			\includegraphics[width=0.15\textwidth]{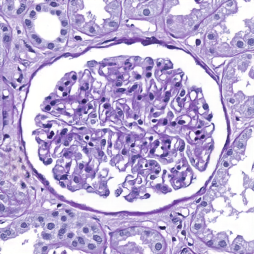}&
			\includegraphics[width=0.15\textwidth]{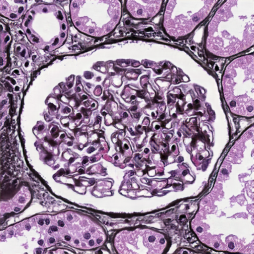} &
			\includegraphics[width=0.15\textwidth]{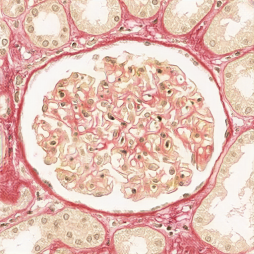} &
			\includegraphics[width=0.15\textwidth]{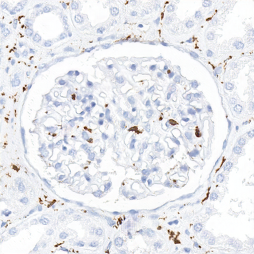} &
			\includegraphics[width=0.15\textwidth]{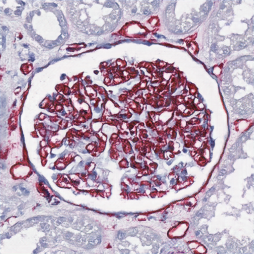}
			\\
			\rowname{Seg.} &
			\includegraphics[width=0.15\textwidth]{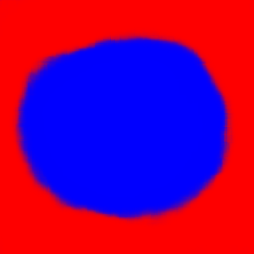}&
			\includegraphics[width=0.15\textwidth]{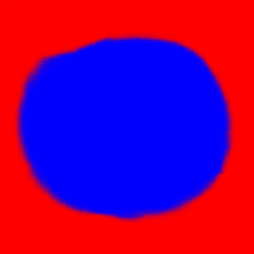}&
			\includegraphics[width=0.15\textwidth]{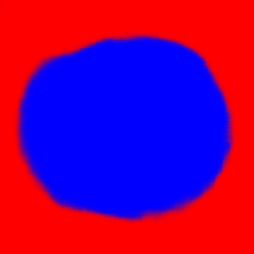} &
			\includegraphics[width=0.15\textwidth]{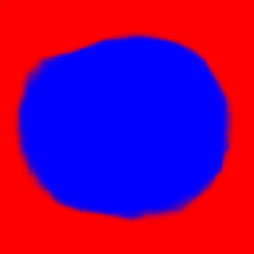} &
			\includegraphics[width=0.15\textwidth]{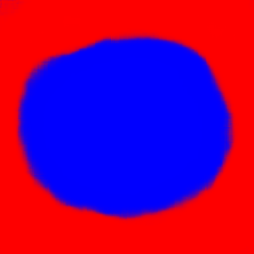} &
			\includegraphics[width=0.15\textwidth]{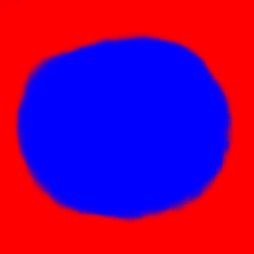}
			\\
			\rowname{CD34} &
			\includegraphics[width=0.15\textwidth]{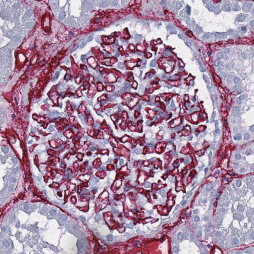}&
			\includegraphics[width=0.15\textwidth]{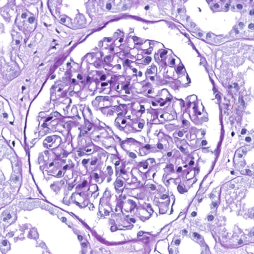}&
			\includegraphics[width=0.15\textwidth]{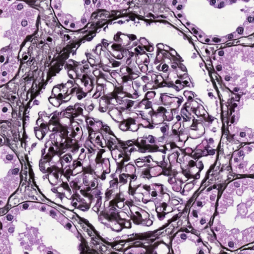} &
			\includegraphics[width=0.15\textwidth]{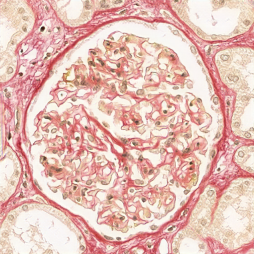} &
			\includegraphics[width=0.15\textwidth]{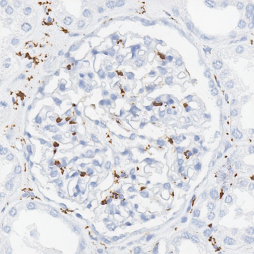} &
			\includegraphics[width=0.15\textwidth]{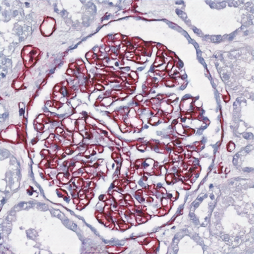}
			\\		
			\rowname{Seg.} &
			\includegraphics[width=0.15\textwidth]{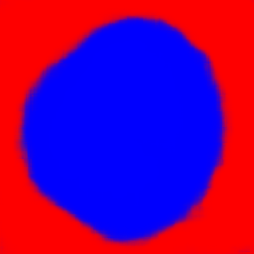}&
			\includegraphics[width=0.15\textwidth]{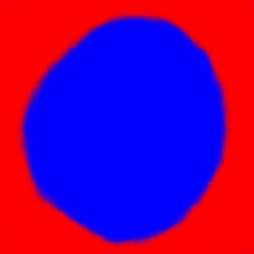}&
			\includegraphics[width=0.15\textwidth]{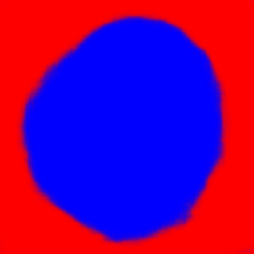} &
			\includegraphics[width=0.15\textwidth]{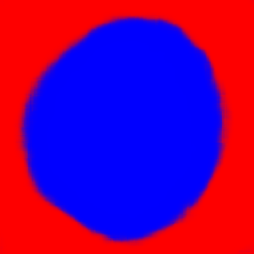} &
			\includegraphics[width=0.15\textwidth]{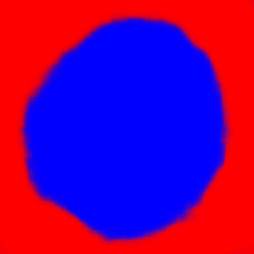} &
			\includegraphics[width=0.15\textwidth]{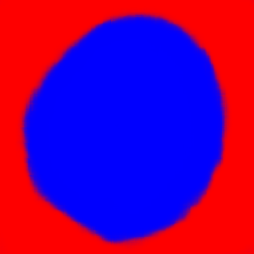}
			\\		
		\end{tabular}
		\caption{HistoStarGAN translations  between different stains with corresponding segmentation. Each translation is obtained by using different latent codes.}
		\label{fig:histostargan_other_translations}
	\end{center}
\end{figure}

A trained model is able to perform diverse stain transfer between any pair of stainings seen during training. The diverse transfer is obtained by sampling different random codes, which are transformed by the mapping network into target-stain specific styles. Some examples of PAS image translations, alongside corresponding segmentations, are provided in Figure \ref{fig:histostargan_pas_translations}. The obtained translations are plausible histopathological images, where the structures of interest, glomeruli in this case, are preserved during translation.  The difference between translations are at the level of microscopic structures (e.g.\ appearance of nuclei, the thickness of a membrane, etc.), see the Figure \ref{fig:histostargan_zppmed_translations}. Since the HistoStarGAN is multi-domain, the same model is also able to perform translations between other staining pairs, examples of which are provided in Figure \ref{fig:histostargan_other_translations}.

\begin{figure}[!htbp]
	\begin{center}
		\settoheight{\tempdima}{\includegraphics[width=0.15\textwidth]{Images_for_HistoStarGANv2/CycleGAN_whole_dataset/03_original.png}}%
		\begin{tabular}{@{}c@{ }c@{ }c@{ }c@{ }c@{ }c@{ }c@{ }c@{ }c@{ }c@{ }}
			& & \multicolumn{5}{c}{Translations}\\
				\cline{3-7} \vspace{-2.5ex}\\
			& Real &PAS &Jones H\&E & Sirius Red & CD68 & CD34\\
			\rowname{H\&E} &
			\includegraphics[width=0.15\textwidth]{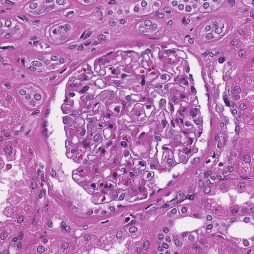}&
			\includegraphics[width=0.15\textwidth]{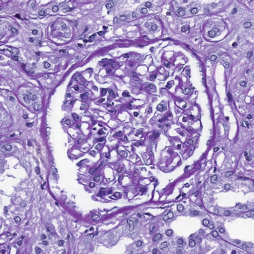}&
			\includegraphics[width=0.15\textwidth]{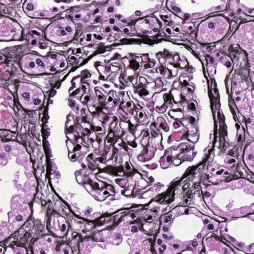} &
			\includegraphics[width=0.15\textwidth]{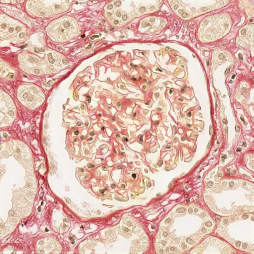} &
			\includegraphics[width=0.15\textwidth]{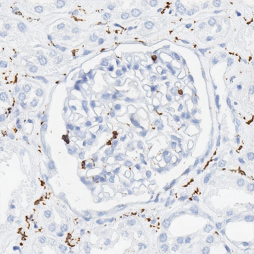} &
			\includegraphics[width=0.15\textwidth]{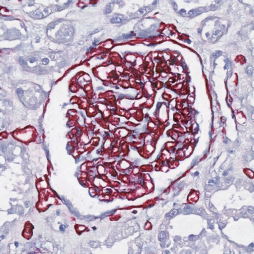}
			\\
			\rowname{Seg.} &
			\includegraphics[width=0.15\textwidth]{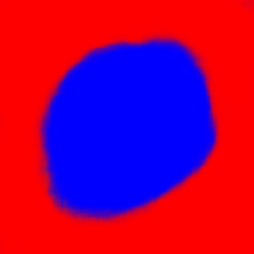}&
			\includegraphics[width=0.15\textwidth]{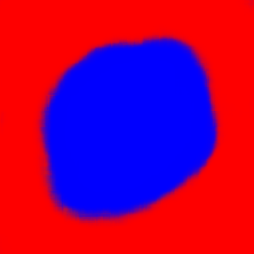}&
			\includegraphics[width=0.15\textwidth]{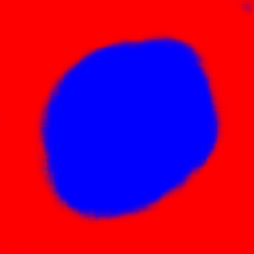} &
			\includegraphics[width=0.15\textwidth]{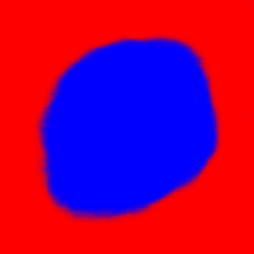} &
			\includegraphics[width=0.15\textwidth]{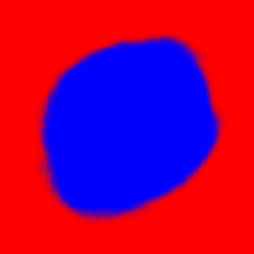} &
			\includegraphics[width=0.15\textwidth]{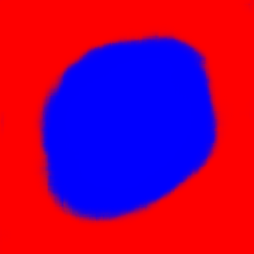}
			\\
			\rowname{CD3-CD68} &
			\includegraphics[width=0.15\textwidth]{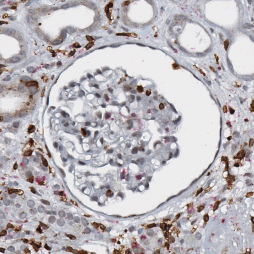}&
			\includegraphics[width=0.15\textwidth]{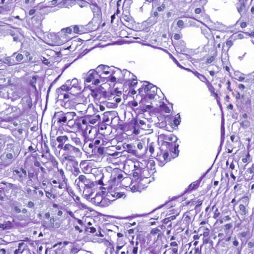}&
			\includegraphics[width=0.15\textwidth]{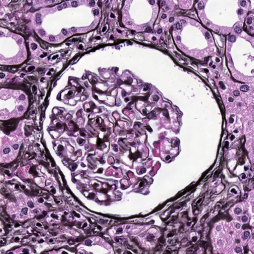} &
			\includegraphics[width=0.15\textwidth]{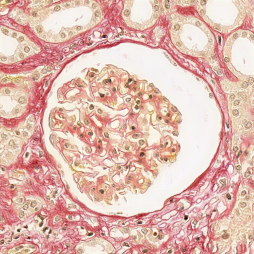} &
			\includegraphics[width=0.15\textwidth]{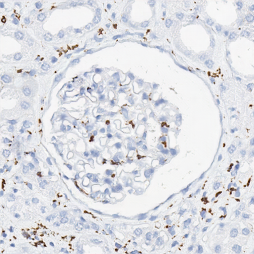} &
			\includegraphics[width=0.15\textwidth]{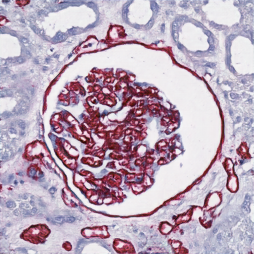}
			\\
			\rowname{Seg.} &
			\includegraphics[width=0.15\textwidth]{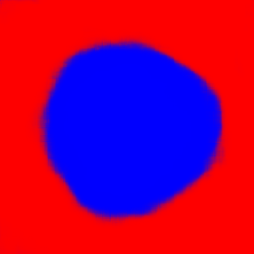}&
			\includegraphics[width=0.15\textwidth]{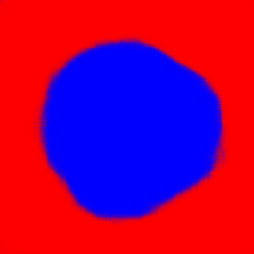}&
			\includegraphics[width=0.15\textwidth]{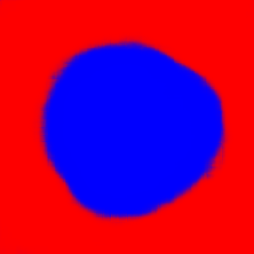} &
			\includegraphics[width=0.15\textwidth]{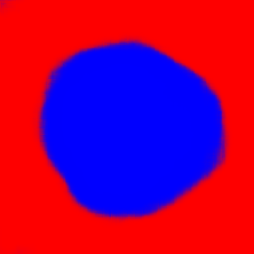} &
			\includegraphics[width=0.15\textwidth]{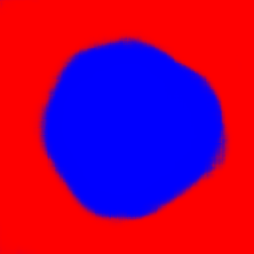} &
			\includegraphics[width=0.15\textwidth]{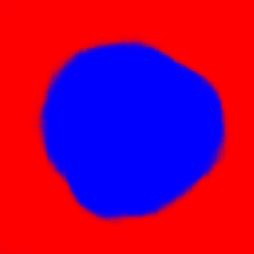}
			\\
			
			\rowname{CD3-CD163} &
			\includegraphics[width=0.15\textwidth]{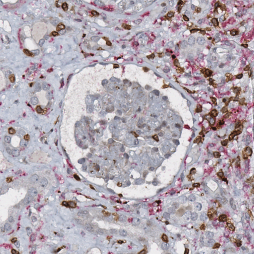}&
			\includegraphics[width=0.15\textwidth]{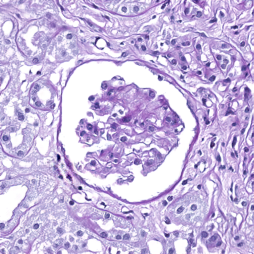}&
			\includegraphics[width=0.15\textwidth]{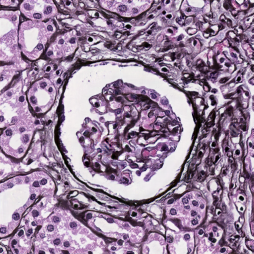} &
			\includegraphics[width=0.15\textwidth]{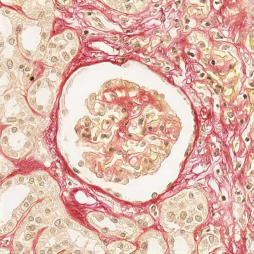} &
			\includegraphics[width=0.15\textwidth]{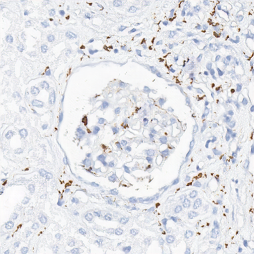} &
			\includegraphics[width=0.15\textwidth]{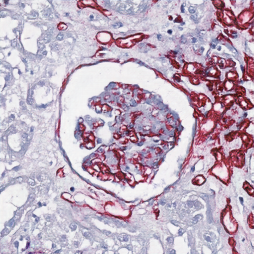}
			\\

			\rowname{Seg.} &
			\includegraphics[width=0.15\textwidth]{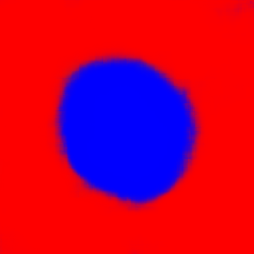}&
			\includegraphics[width=0.15\textwidth]{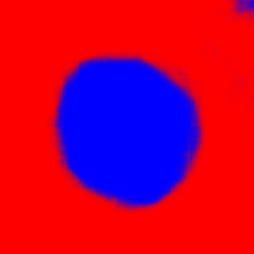}&
			\includegraphics[width=0.15\textwidth]{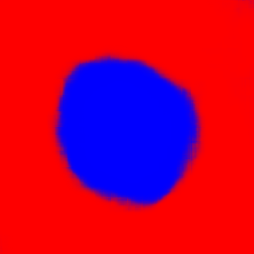} &
			\includegraphics[width=0.15\textwidth]{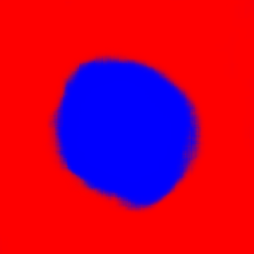} &
			\includegraphics[width=0.15\textwidth]{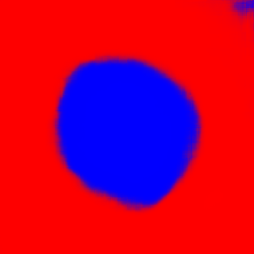} &
			\includegraphics[width=0.15\textwidth]{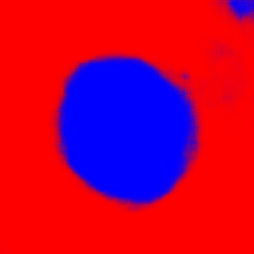}
			\\
			
			\rowname{CD3-CD206} &
			\includegraphics[width=0.15\textwidth]{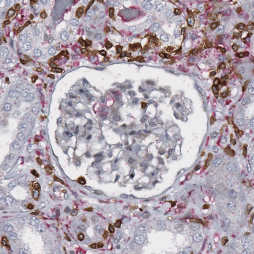}&
			\includegraphics[width=0.15\textwidth]{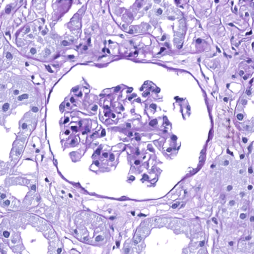}&
			\includegraphics[width=0.15\textwidth]{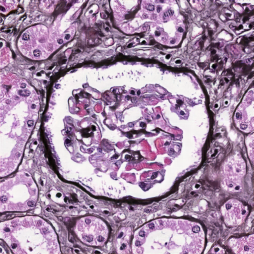} &
			\includegraphics[width=0.15\textwidth]{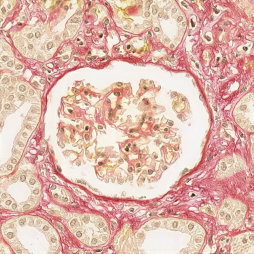} &
			\includegraphics[width=0.15\textwidth]{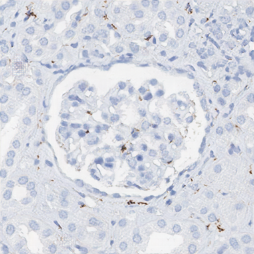} &
			\includegraphics[width=0.15\textwidth]{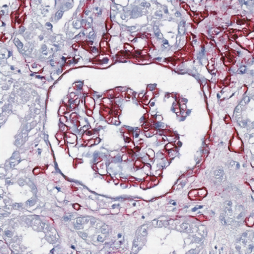}
			\\			
			\rowname{Seg.} &
			\includegraphics[width=0.15\textwidth]{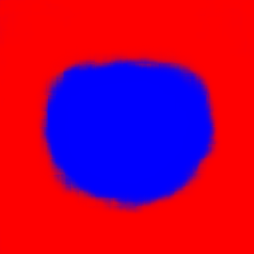}&
			\includegraphics[width=0.15\textwidth]{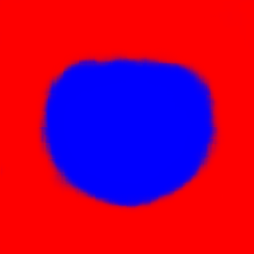}&
			\includegraphics[width=0.15\textwidth]{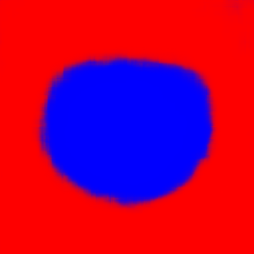} &
			\includegraphics[width=0.15\textwidth]{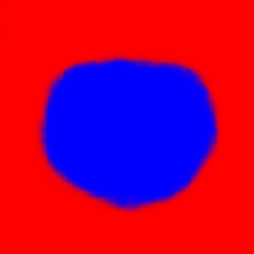} &
			\includegraphics[width=0.15\textwidth]{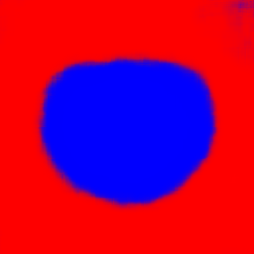} &
			\includegraphics[width=0.15\textwidth]{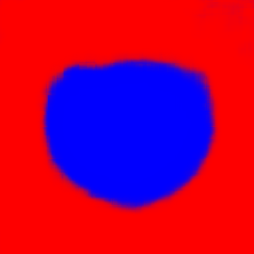}
		\end{tabular}
		\caption{HistoStarGAN -- generalisation of stain transfer and segmentation to unseen stain modalities.}
		\label{fig:histostargan_unseen_translations}
	\end{center}
\end{figure}

\begin{figure}[!htbp]
	\begin{center}
		\settoheight{\tempdima}{\includegraphics[width=0.15\textwidth]{Images_for_HistoStarGANv2/CycleGAN_whole_dataset/03_original.png}}%
		\begin{tabular}{@{}c@{ }c@{ }c@{ }c@{ }c@{ }c@{ }c@{ }c@{ }c@{ }c@{ }}
			& & \multicolumn{5}{c}{Translations}\\
			\cline{3-7} \vspace{-1.5ex}\\
			& Real &PAS &Jones H\&E & Sirius Red & CD68 & CD34\\
			\rowname{AIDPATH 1} &
			\includegraphics[width=0.15\textwidth]{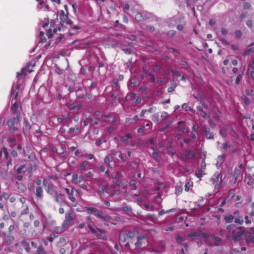}&
			\includegraphics[width=0.15\textwidth]{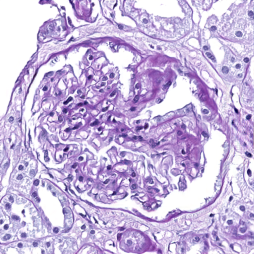}&
			\includegraphics[width=0.15\textwidth]{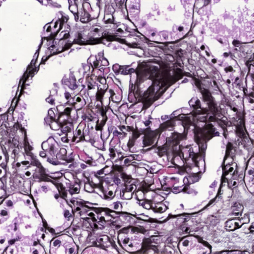} &
			\includegraphics[width=0.15\textwidth]{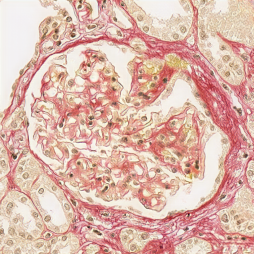} &
			\includegraphics[width=0.15\textwidth]{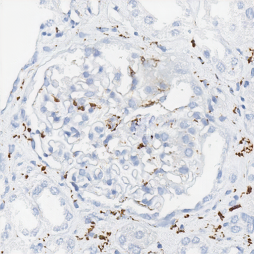} &
			\includegraphics[width=0.15\textwidth]{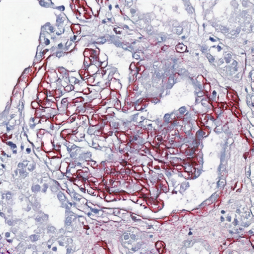}
			\\
			\rowname{Seg.} &
			\includegraphics[width=0.15\textwidth]{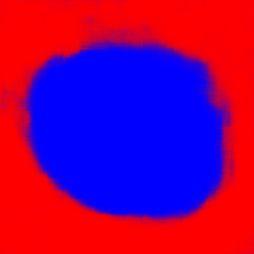}&
			\includegraphics[width=0.15\textwidth]{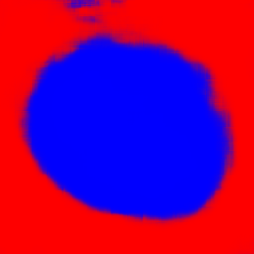}&
			\includegraphics[width=0.15\textwidth]{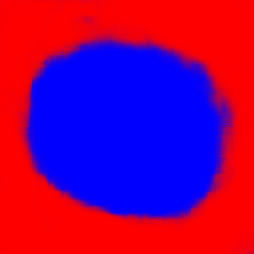} &
			\includegraphics[width=0.15\textwidth]{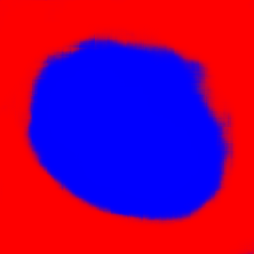} &
			\includegraphics[width=0.15\textwidth]{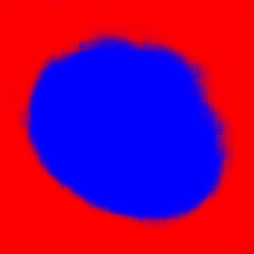} &
			\includegraphics[width=0.15\textwidth]{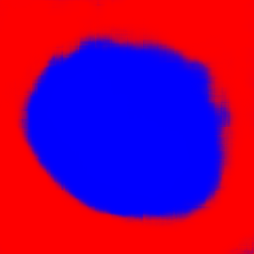}
			\\
			\rowname{AIDPATH 2} &
			\includegraphics[width=0.15\textwidth]{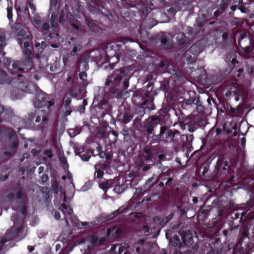}&
			\includegraphics[width=0.15\textwidth]{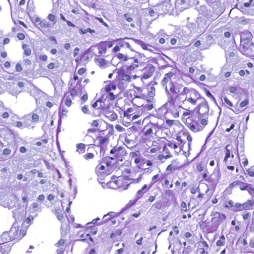}&
			\includegraphics[width=0.15\textwidth]{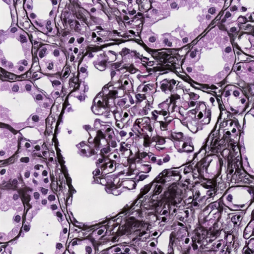} &
			\includegraphics[width=0.15\textwidth]{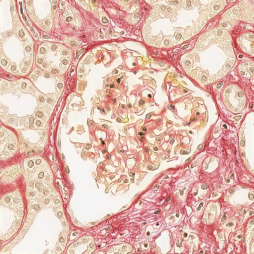} &
			\includegraphics[width=0.15\textwidth]{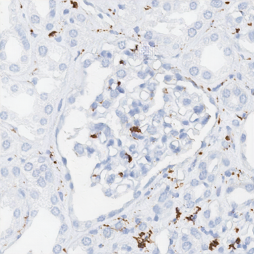} &
			\includegraphics[width=0.15\textwidth]{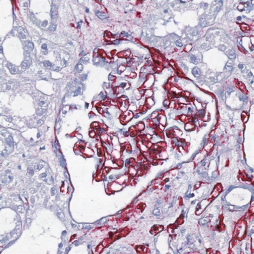}
			\\
			\rowname{Seg.} &
			\includegraphics[width=0.15\textwidth]{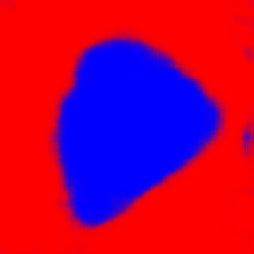}&
			\includegraphics[width=0.15\textwidth]{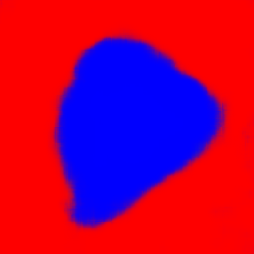}&
			\includegraphics[width=0.15\textwidth]{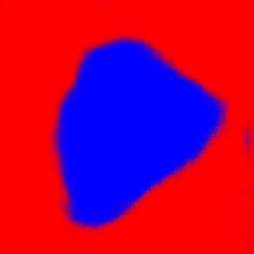} &
			\includegraphics[width=0.15\textwidth]{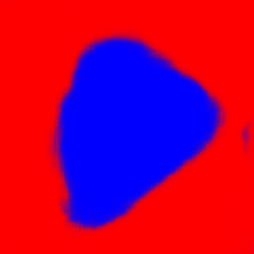} &
			\includegraphics[width=0.15\textwidth]{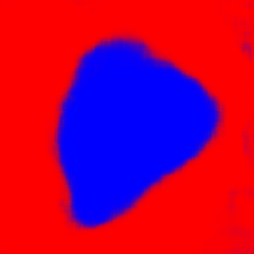} &
			\includegraphics[width=0.15\textwidth]{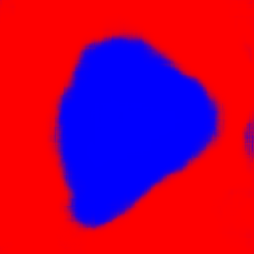}
		\end{tabular}
		\caption{HistoStarGAN applied for stain normalisation, stain transfer and glomeruli segmentation of the publicly available AIDPATH (PAS-based) dataset.}
		\label{fig:histostargan_aidpath_translations}
	\end{center}
\end{figure}

\textbf{Generalisation of stain transfer:} Since  HistoStarGAN is trained on a variety of stains, the generator's encoder is stain invariant, which for the first time enables virtual staining of unseen stains. Examples of which are presented in Figure \ref{fig:histostargan_unseen_translations}, in which a new stain modality named H\&E in addition to three double-stainings CD3-CD68, CD3-CD163 and CD3-CD206 are translated to stainings seen during training. 
Moreover, in Figure \ref{fig:histostargan_aidpath_translations} the model is applied to the AIDPATH dataset \cite{aidpath_dataset} composed of images which are publicly available variations of the PAS stain. This demonstrates that HistoStarGAN can generalise and obtain stain normalisation (column PAS) and stain transfer (other columns) simultaneously, alongside stain-invariant segmentation. Videos representing the exploration of the latent space during translation, are provided online\footnote{\href{https://main.d33ezaxrmu3m4a.amplifyapp.com/}{ https://main.d33ezaxrmu3m4a.amplifyapp.com/}}. 

\begin{table}[!htb]
	\centering
	\small{
	\begin{tabularx}{\textwidth}{L{1.6cm} X X X X X X X}
			\multirow{2}{=}{\shortstack{\textbf{Model}}} & \multirow{2}{=}{\textbf{Score}} & \multicolumn{5}{c}{\textbf{Test Staining}}\\
			& & PAS & Jones H\&E & CD68 & Sirius Red & CD34 & Overall\\
			\hhline{========}
			\multirow{3}{=}{{UDA-GAN}} & F$_1$&\textbf{0.903} \footnotesize{(0.003)} & 0.849 \footnotesize{(0.031)} & 0.720 \footnotesize{(0.016)} & \textbf{0.875} \footnotesize{(0.016)} & 0.800 \footnotesize{(0.033)} &  0.829 \footnotesize{(0.072)}\\
			& Precision & 0.878 \footnotesize{(0.018)} & 0.787 \footnotesize{(0.060)} & 0.688 \footnotesize{(0.110)} &0.835 \footnotesize{(0.035)} & 0.720 \footnotesize{(0.064)} &  0.782 \footnotesize{(0.079)}\\ 
			& Recall & 0.930 \footnotesize{(0.014)} & 0.923 \footnotesize{(0.010)} & 0.777 \footnotesize{(0.095)} &0.921 \footnotesize{(0.007)} &0.903 \footnotesize{(0.016)} &  0.891 \footnotesize{(0.064)} \\
			\midrule
			\multirow{3}{=}{{HistoStar-GAN}} & F$_1$ & 0.871 \footnotesize{(0.009)}  & \textbf{0.870} \footnotesize{(0.007)} & \textbf{0.755} \footnotesize{(0.006)} & 0.859 \footnotesize{(0.004)} &\textbf{0.840 }\footnotesize{(0.004)}&\textbf{0.839} \footnotesize{(0.048)} \\
			& Precision & 0.845 \footnotesize{(0.029)} & 0.864 \footnotesize{(0.019)} &0.845 \footnotesize{(0.039)} &0.883 \footnotesize{(0.018)} &0.839 \footnotesize{(0.032)} & 0.855 \footnotesize{(0.018)} \\ 
			& Recall & 0.899 \footnotesize{(0.016)} & 0.877 \footnotesize{(0.007)} & 0.684 \footnotesize{(0.024)} & 0.836 \footnotesize{(0.017)} &0.842 \footnotesize{(0.024)}& 0.828 \footnotesize{(0.084)}\\ 

		\end{tabularx}
	}
	\caption{ Quantitative results for HistoStarGAN compared to UDA-GAN. Each model is trained on annotated PAS (source staining) and tested on different (target) stainings. Standard deviations are in parentheses, the highest F$_1$ scores for each staining are in bold.}
	\label{tab: histostargan_quantitative results}
\end{table}
\begin{table}[!htb]
	\centering
	\small{
		\begin{tabularx}{\textwidth}{L{1.6cm} L{1.5cm} X X X X X X X}
			\multirow{3}{*}{\shortstack{{\textbf{Model}}}} & \multirow{3}{*}{{\textbf{Score}}} & \multicolumn{6}{c}{\textbf{{Test Staining}}}\\
			& & H\&E & CD3 & CD3-CD68 & CD3-CD163 & CD3-CD206 & CD3-MS4A4A &Overall\\
			\hhline{=========}
			\multirow{5}{=}{UDA-GAN} & F$_1$ &0.681 \footnotesize{(0.031)} & 0.648 \footnotesize{(0.111)} & 0.258 \footnotesize{(0.062} & 0.260 \footnotesize{(0.050)} & 0.330 \footnotesize{(0.058)} &  0.330 \footnotesize{(0.071)} & 0.418 \footnotesize{(0.194)}\\
			& Precision & 0.865 \footnotesize{(0.079)} & 0.550 \footnotesize{(0.161)} & 0.171 \footnotesize{(0.047)} &0.168 \footnotesize{(0.039)} & 0.230 \footnotesize{(0.054)} &  0.240 \footnotesize{(0.070)}& 0.371 \footnotesize{(0.281)}\\ 
			& Recall & 0.563 \footnotesize{(0.028)} & 0.824 \footnotesize{(0.032)} & 0.538 \footnotesize{(0.070)} &0.586 \footnotesize{(0.039)} &0.598 \footnotesize{(0.029)} &  0.542 \footnotesize{(0.029)} & 0.608 \footnotesize{(0.108)}\\
			\midrule
			\multirow{5}{=}{HistoStar-GAN} & F$_1$ &\textbf{0.813} \footnotesize{(0.022)}  & \textbf{0.741} \footnotesize{(0.009)} & \textbf{0.597} \footnotesize{(0.011)} & \textbf{0.611} \footnotesize{(0.015)} &\textbf{0.593 }\footnotesize{(0.014)}&\textbf{0.570} \footnotesize{(0.012)} &\textbf{0.654} \footnotesize{(0.099)}\\
			& Precision & 0.855 \footnotesize{(0.018)} & 0.850 \footnotesize{(0.007)} &0.835 \footnotesize{(0.022)} &0.891 \footnotesize{(0.021)} &0.881 \footnotesize{(0.026)} & 0.882 \footnotesize{(0.025)} & 0.866 \footnotesize{(0.022)}\\
			& Recall & 0.777 \footnotesize{(0.056)} & 0.656 \footnotesize{(0.011)} & 0.465 \footnotesize{(0.017)} & 0.465 \footnotesize{(0.021)} & 0.447 \footnotesize{(0.020)}& 0.422 \footnotesize{(0.016)}& 0.539 \footnotesize{(0.144)}\\
			
		\end{tabularx}
	}
	\caption{ Quantitative results for HistoStarGAN compared to UDA-GAN on unseen stains. Each model is trained on annotated PAS (source staining). Standard deviations are in parentheses, the highest F$_1$ scores for each staining are in bold.}
	\label{tab: histostargan_quantitative results_unseen_stains}
\end{table}

A model composed of the generator's encoder and the segmentation branch, can perform stain-invariant segmentation of WSIs across various stainings. Table \ref{tab: histostargan_quantitative results} presents  the segmentation results for test WSIs from all stainings (virtually) seen during training. The model's performance is compared to UDA-GAN, which uses the same CycleGAN models for data augmentation. Since the patch size is $512 \times 512$, each patch is cropped to $508\times 508$ during UDA-GAN training. The presented results are the averages of three independent training repetitions with corresponding standard deviations.

The HistoStarGAN can generalise across all virtually seen stainings during training,  outperforming UDA-GAN trained using the same translation models. Apart from being an end-to-end model that simultaneously performs virtual staining and segmentation,   HistoStarGAN also results in an increase in precision. 
This can be attributed to the fact that HistoStarGAN better recognises the negative tissue (less false positives). However, recall is lower, which indicates that more glomeruli (or parts of them) are missed compared to UDA-GAN. 
This could be the consequence of predicting a segmentation mask from the feature space directly, without skip-connections.  However, extending the HistoStarGAN model with skip-connections between the encoder and segmentation branch, experimentally showed to  negatively affect training stability.

\textbf{Generalisation of stain invariance:} To test the generalisation of HistoStarGAN, the model is applied to new stainings not seen during training.  Using the annotations of four  whole-slide images from each of the new stainings (H\&E,  CD3,  CD3-CD68, CD3-CD163 and CD3-CD206), the averages of three independent training repetitions with corresponding standard deviations are presented in Table \ref{tab: histostargan_quantitative results_unseen_stains}. Overall,  HistoStarGAN generalises better and is more robust compared to UDA-GAN. A potential cause of UDA-GAN failure in some stains can be its learning process where stain invariance is forced only on the pixel level. Although the HistoStarGAN uses the identical CyleGAN translations, the framework extracts stain-invariant features, which are better transferred to unseen stains. Moreover, the segmentation branch of HistoStarGAN is trained on a dataset with more variety since HistoStarGAN translations are also included (see Figure \ref{fig:starganv2_diagram}).

\section{Ablation Studies}
\label{sec:chapter_histostargan:ablation_study}
\begin{figure}[!htbp] 
	\centering 
	\includegraphics[width=0.3\textwidth]{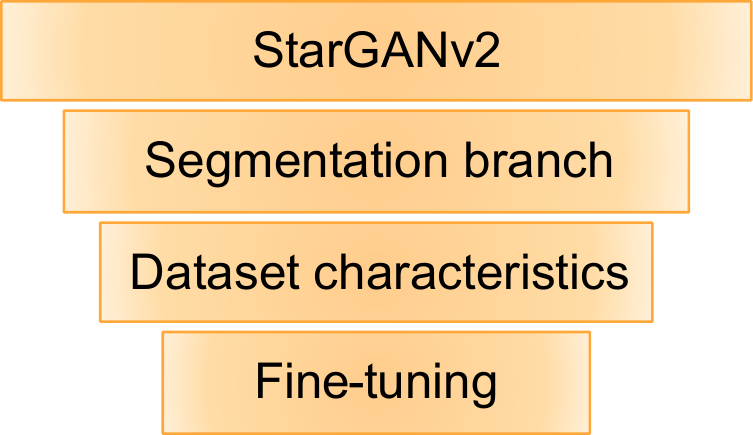}
	\caption[HistoStarGAN - Illustration of ablation studies]{Illustration of the differences between the StarGANv2 and HistoStarGAN. Each of these differences is justified via an ablation study.}
	\label{fig:histostargan_ablation_diagram}
\end{figure}

HistoStarGAN model  builds upon StarGANv2 in several aspects -- the first extension is attaching a segmentation module to the generator; the second is using  pre-trained CycleGANs to create  the training dataset, and the third is fine-tuning the segmentation module using only source data. In the following, each of these aspects will be discussed via an ablation study, in a bottom-to-top direction as illustrated in  Figure \ref{fig:histostargan_ablation_diagram}.

\begin{table}[!htb]
	\centering
	\small{
		\begin{tabularx}{\textwidth}{L{1.9cm} L{1.5cm} X X X X X X}
			\multirow{2}{=}{\shortstack{{\textbf{Model}}}} & \multirow{2}{=}{{\textbf{Score}}} & \multicolumn{6}{c}{{\textbf{Test Staining}}}\\
			& & PAS & Jones H\&E & CD68 & Sirius Red & CD34 & Overall \\
			\hhline{========}\\
			\multirow{3}{=}{HistoStar-GAN} & F$_1$ &0.820 & 0.816  & 0.705 & 0.792  & 0.777 &  0.782 \\
			& Precision  &0.779 & 0.791  & 0.764 & 0.792  & 0.802 &  0.761 \\ 
			& Recall & 0.865  & 0.843 & 0.655  & 0.782 &  0.795 & 0.779 \\
			\midrule
			\multirow{3}{=}{{HistoStar-GAN (fine-tuned)}} & F$_1$ &\textbf{0.876} & \textbf{0.875}  & \textbf{0.762} & \textbf{0.855}  & \textbf{0.842} &  \textbf{0.842} \\
			& Precision  &\textbf{0.851}&\textbf{0.871}&\textbf{0.861}& \textbf{0.886}&\textbf{0.842}&\textbf{0.862}\\ 
			& Recall &\textbf{0.902}&\textbf{0.879} & \textbf{0.683}&\textbf{0.825} &\textbf{0.843} &\textbf{0.826}\\
		\end{tabularx}
	}
	\caption{ Fine-tuning effects on the  segmentation model's performance, in which the model is fine-tuned for one epoch. }
	\label{tab: histostargan_finetuning_effect}
\end{table}

\subsection{Fine-Tuning}

Table \ref{tab: histostargan_finetuning_effect} demonstrates that fine-tuning the segmentation branch using the fixed generator's encoder increases segmentation performance. In practice, fine-tuning for just a single epoch on an imbalanced source dataset increases overall performance on virtually seen stains (test patient WSIs) by around $6\%$ in overall $F_1$-score. However, longer fine-tuning, although potentially beneficial for particular stainings, does not offer any benefits. Also, the performance on the source's validation set does not correlate with the obtained improvements, i.e.\ the fine-tuned model with the lowest validation loss does not bring the best overall results on unseen stains. Thus, fine-tuning for one epoch is chosen.

\subsection{Dataset Characteristics}
A balanced and (virtually) fully annotated dataset is used to train HistoStarGAN.  A labelled dataset is required since the segmentation branch is trained in a supervised manner, and thus alternative data sampling strategies such as random data sampling \cite{gadermayr2018which,udagan} are not suitable. However, the fully annotated dataset does not need to be balanced. Thus,  using the findings of \citet{lampert2019strategies}, an imbalanced source (PAS) dataset is formed where all glomeruli in the training patient images are extracted ($662$) and seven times as many negative patches ($4634$). This dataset is translated using CycleGAN to all target stains (CycleGAN models are always trained in an unsupervised way, on random patches). Thus, a fully annotated and highly imbalanced dataset is created on top of which the HistoStarGAN model is trained.

\begin{figure}[!htbp]
	\begin{center}
		\settoheight{\tempdima}{\includegraphics[width=0.15\textwidth]{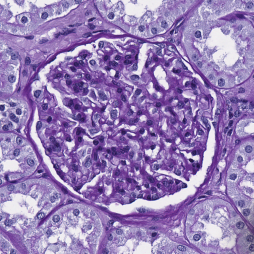}}%
		\begin{tabular}{@{}c@{ }c@{ }c@{ }c@{ }c@{ }c@{ }c@{ }c@{ }c@{ }c@{ }}
			& \multicolumn{5}{c}{Translations}\\
			\cline{2-6} \vspace{-1.5ex}\\
			Real PAS &PAS &Jones H\&E & Sirius Red & CD68 & CD34\\
			\includegraphics[width=0.15\textwidth]{Images_for_HistoStarGANv2/HistoStarGAN_whole_dataset_w_5/02_original.png}&
			\includegraphics[width=0.15\textwidth]{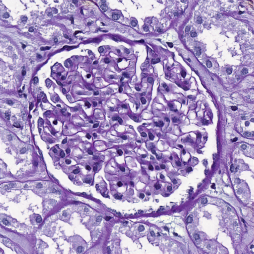}&
			\includegraphics[width=0.15\textwidth]{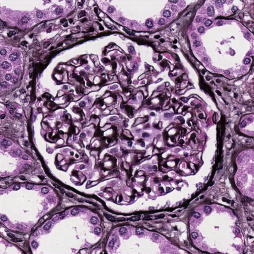} &
			\includegraphics[width=0.15\textwidth]{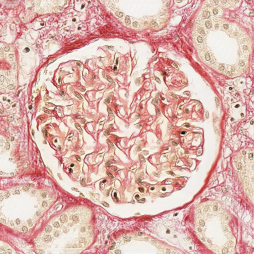} &
			\includegraphics[width=0.15\textwidth]{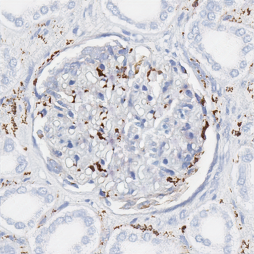} &
			\includegraphics[width=0.15\textwidth]{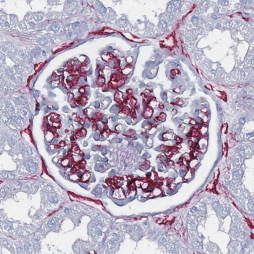}
			\\
			
			\includegraphics[width=0.15\textwidth]{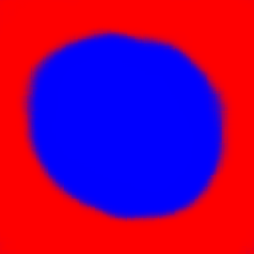}&
			\includegraphics[width=0.15\textwidth]{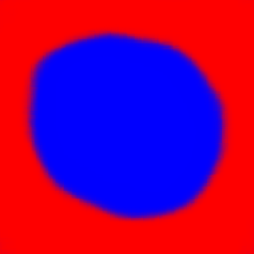}&
			\includegraphics[width=0.15\textwidth]{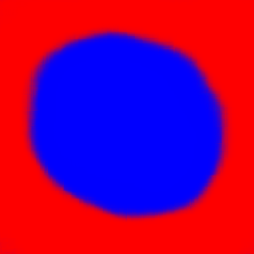} &
			\includegraphics[width=0.15\textwidth]{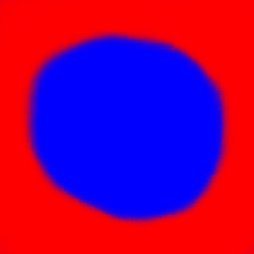} &
			\includegraphics[width=0.15\textwidth]{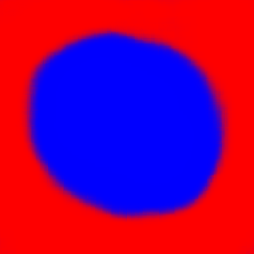} &
			\includegraphics[width=0.15\textwidth]{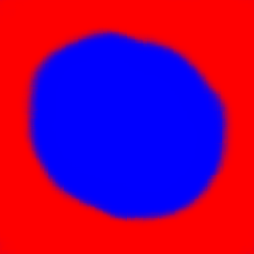}
			\\
			&
			\includegraphics[width=0.15\textwidth]{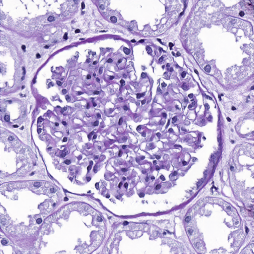}&
			\includegraphics[width=0.15\textwidth]{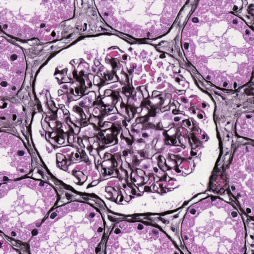} &
			\includegraphics[width=0.15\textwidth]{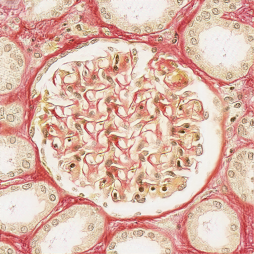} &
			\includegraphics[width=0.15\textwidth]{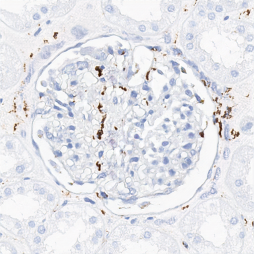} &
			\includegraphics[width=0.15\textwidth]{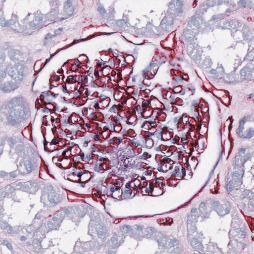}
			\\
			&
			\includegraphics[width=0.15\textwidth]{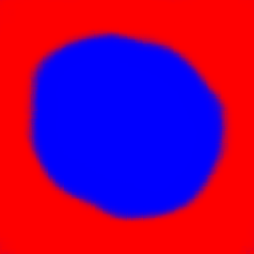}&
			\includegraphics[width=0.15\textwidth]{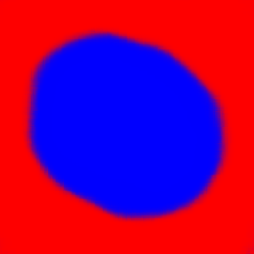} &
			\includegraphics[width=0.15\textwidth]{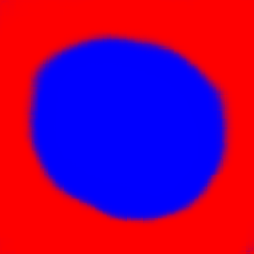} &
			\includegraphics[width=0.15\textwidth]{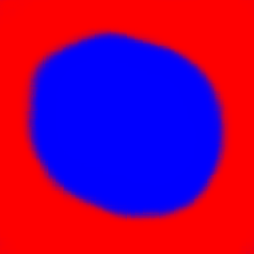} &
			\includegraphics[width=0.15\textwidth]{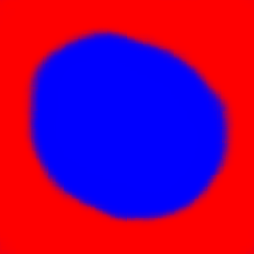}
			\\
			&
			\includegraphics[width=0.15\textwidth]{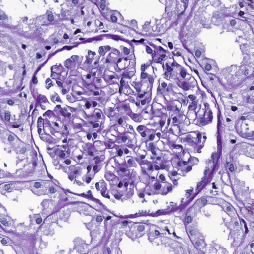}&
			\includegraphics[width=0.15\textwidth]{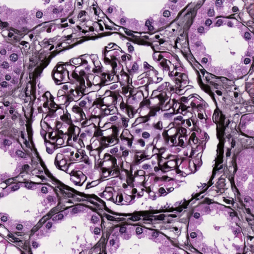} &
			\includegraphics[width=0.15\textwidth]{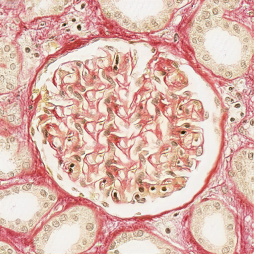} &
			\includegraphics[width=0.15\textwidth]{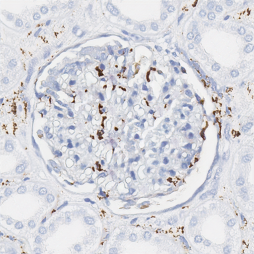} &
			\includegraphics[width=0.15\textwidth]{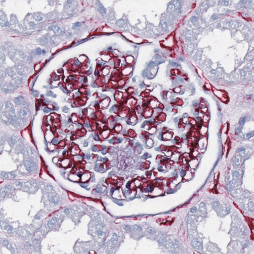}
			\\
			&
			\includegraphics[width=0.15\textwidth]{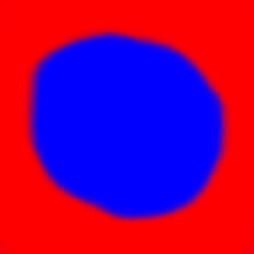}&
			\includegraphics[width=0.15\textwidth]{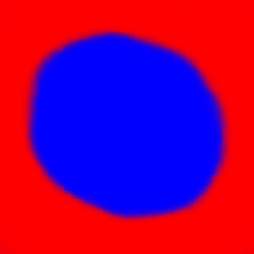} &						\includegraphics[width=0.15\textwidth]{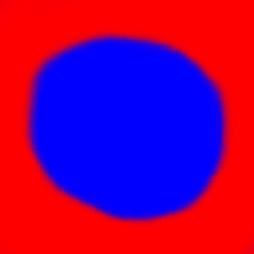} &
			\includegraphics[width=0.15\textwidth]{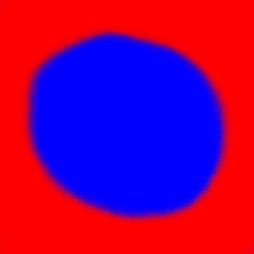} &
			\includegraphics[width=0.15\textwidth]{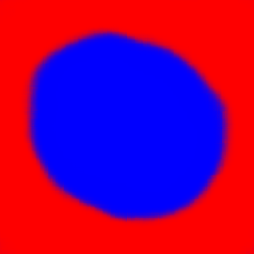}
			\\
		\end{tabular}
		\caption[HistoStarGAN trained on imbalanced dataset]{ HistoStarGAN  translations using a model trained using a CycleGAN-generated imbalanced annotated dataset. The images in each column are generated using the same latent codes.}
		\label{fig:histostarganv2_cyclegan_whole_dataset_w_5}
	\end{center}
\end{figure}

The results obtained in this setting are presented in Figure \ref{fig:histostarganv2_cyclegan_whole_dataset_w_5}, in which one glomeruli patch from the PAS stain is translated into multiple stainings using different latent codes. The obtained translations for one stain pair differ from each other in terms of glomeruli texture and surrounding structures. The model is able to change the size of glomeruli by changing the size of the Bowman's capsule (white space), in addition to varying the appearance of stain-specific markers (i.e.\ macrophages in CD68 stain). Nevertheless, the segmentation branch successfully recognises all of glomeruli variations.  Compared to training with a balanced dataset, this setting offers more translation variability related to the surrounding tissue and structure of the glomeruli themselves. However, although globally these translations look realistic,  the internal structure inside the glomeruli are not.  Since in HistoStarGAN these translations must also be successfully segmented, this leads to an increase in the false positive rate. For example, it is evident that the produced translations often contain `artificial looking patterns' such as a tendency to group microscopic structures (e.g.\ nuclei) into diagonal, horizontal or vertical stripes (in Figure \ref{fig:histostarganv2_cyclegan_whole_dataset_w_5} this is mostly visible in translations to Sirius Red and Jones H\&E). Therefore, the proposed model uses a balanced dataset since it greatly reduces such possibilities.
 
\subsection{Segmentation Branch}
\label{sec:ablation_study_segmenatation_branch}
Without the segmentation branch, HistoStarGAN  is reduced  to StarGANv2 (see Figure \ref{fig:histostargan_ablation_diagram}). Moreover, removing the segmentation module removes the requirement for the dataset to be fully annotated. Thus, such a model can be trained using a dataset composed of random patches (uniformly sampled in each stain), as well as created by CycleGAN translations of the annotated stain (PAS), both balanced and imbalanced. Each of these will be separately analysed. 

\begin{figure}[!htbp]
	\begin{center}
		\settoheight{\tempdima}{\includegraphics[width=0.145\textwidth]{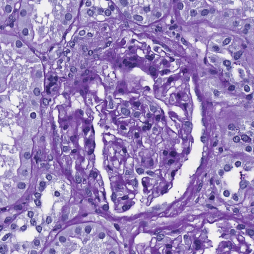}}%
		\begin{tabular}{@{}c@{ }c@{ }c@{ }c@{ }c@{ }c@{ }c@{ }c@{ }c@{ }c@{ }c@{ }}
			& &\multicolumn{5}{c}{Translations}\\
			\cline{3-7}\vspace{-1.5ex} \\
			& Real image &PAS &Jones H\&E & Sirius Red & CD68 & CD34\\
			\rowname{PAS} &
			\includegraphics[width=0.145\textwidth]{Images_for_HistoStarGANv2/CycleGAN/02_orig.png}&
			\includegraphics[width=0.145\textwidth]{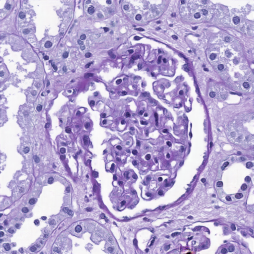}&
			\includegraphics[width=0.145\textwidth]{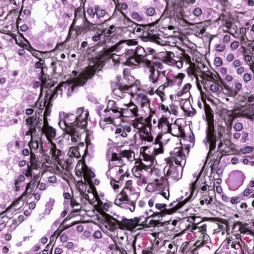} &
			\includegraphics[width=0.145\textwidth]{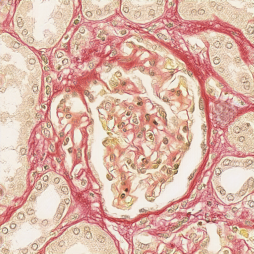} &
			\includegraphics[width=0.145\textwidth]{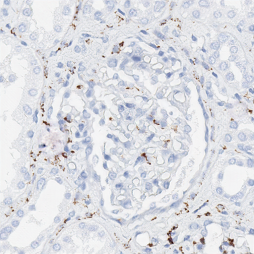} &
			\includegraphics[width=0.145\textwidth]{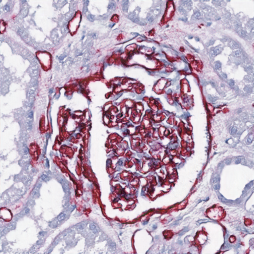}
			\\
			\rowname{Jones H\&E} &
			\includegraphics[width=0.145\textwidth]{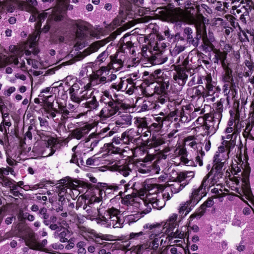}&
			\includegraphics[width=0.145\textwidth]{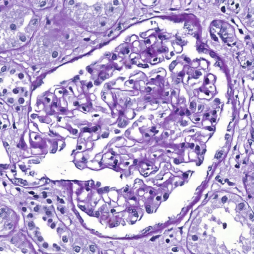}&
			\includegraphics[width=0.145\textwidth]{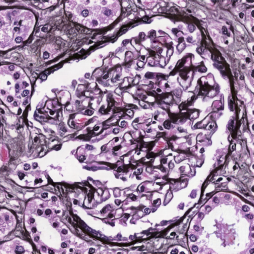} &
			\includegraphics[width=0.145\textwidth]{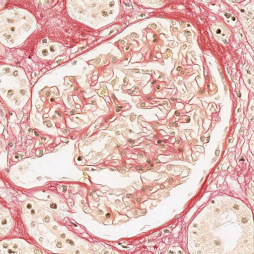} &
			\includegraphics[width=0.145\textwidth]{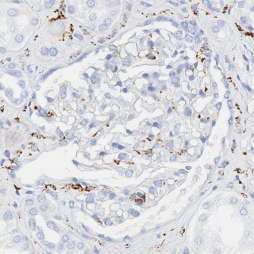} &
			\includegraphics[width=0.145\textwidth]{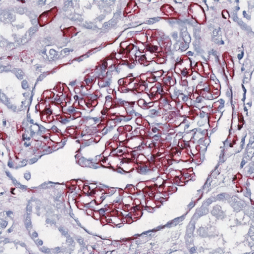}
			\\
			\rowname{CD68} &	
			\includegraphics[width=0.145\textwidth]{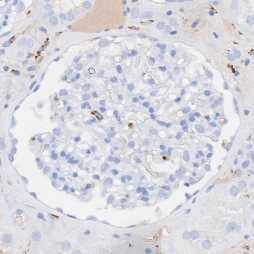}&
			\includegraphics[width=0.145\textwidth]{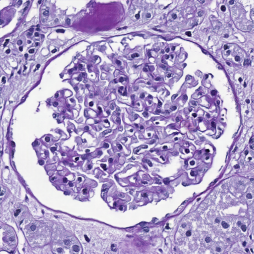}&
			\includegraphics[width=0.145\textwidth]{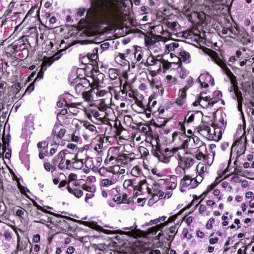} &
			\includegraphics[width=0.145\textwidth]{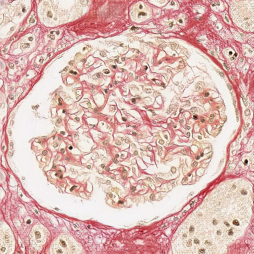} &
			\includegraphics[width=0.145\textwidth]{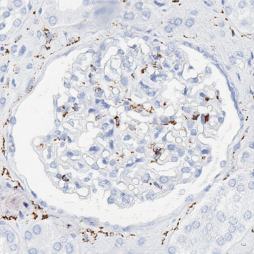} &
			\includegraphics[width=0.145\textwidth]{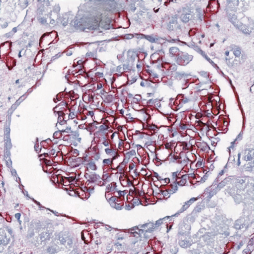}
			\\
			\rowname{Sirius Red} &	
			\includegraphics[width=0.145\textwidth]{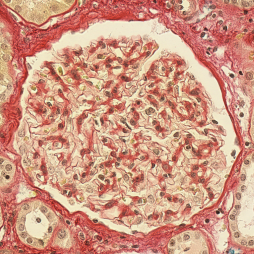}&
			\includegraphics[width=0.145\textwidth]{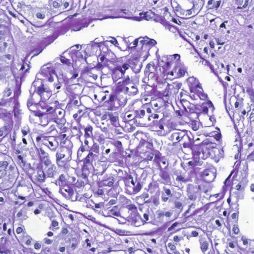}&
			\includegraphics[width=0.145\textwidth]{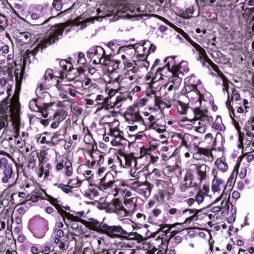} &
			\includegraphics[width=0.145\textwidth]{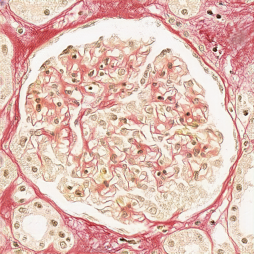} &
			\includegraphics[width=0.145\textwidth]{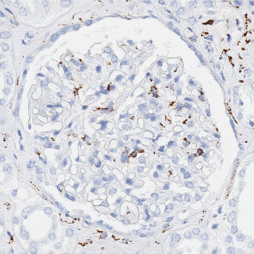} &
			\includegraphics[width=0.145\textwidth]{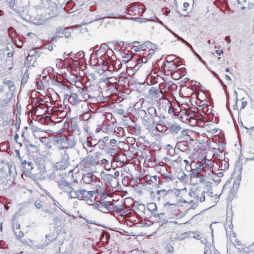}
			\\
			\rowname{CD34} &		
			\includegraphics[width=0.145\textwidth]{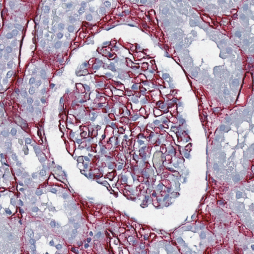}&
			\includegraphics[width=0.145\textwidth]{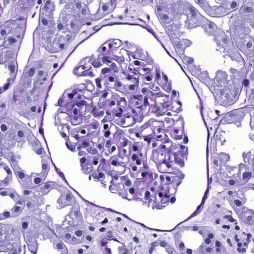}&
			\includegraphics[width=0.145\textwidth]{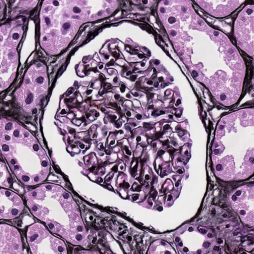} &
			\includegraphics[width=0.145\textwidth]{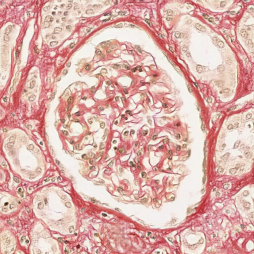} &
			\includegraphics[width=0.145\textwidth]{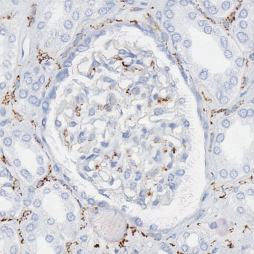} &
			\includegraphics[width=0.145\textwidth]{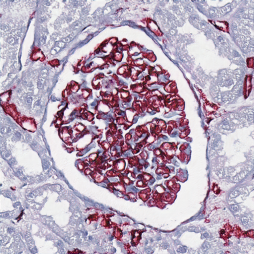}
			\\

		\end{tabular}
		\caption[StarGANv2 trained on balanced dataset]{ StarGANv2 trained on CycleGAN-generated translations of a balanced PAS dataset to other stains. Each image in generated using different latent code.}
		\label{fig:starganv2_cyclegan_dataset_other_translation}
	\end{center}
\end{figure}

\begin{figure}[!htbp]
	\begin{center}
		\settoheight{\tempdima}{\includegraphics[width=0.14\textwidth]{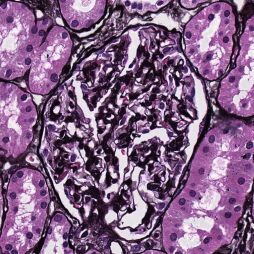}}%
		\begin{tabular}{@{}c@{ }c@{ }c@{ }c@{ }c@{ }c@{ }c@{ }c@{ }c@{ }c@{ }c@{ }}
			& & &\multicolumn{5}{c}{Translation}\\
			\cline{4-8}\vspace{-1.5ex} \\
			& & Real Patch &PAS &Jones H\&E &  Sirius Red & CD68  & CD34\\
			\multirow{2}{*}{\rowname{StarGANv2}} & \rowname{}&
			\includegraphics[width=0.14\textwidth]{Images_for_HistoStarGANv2/avg_translations/avg_mds1_images_for_latex/avg_mds1_50_IFTA_Nx_0010_03_glomeruli_patch_35_orig.png}&
			\includegraphics[width=0.14\textwidth]{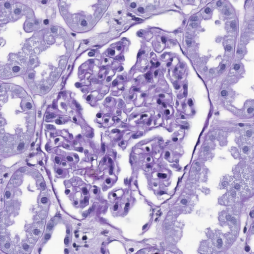}&
			\includegraphics[width=0.14\textwidth]{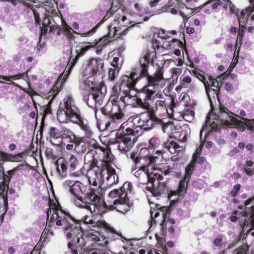}&
			\includegraphics[width=0.14\textwidth]{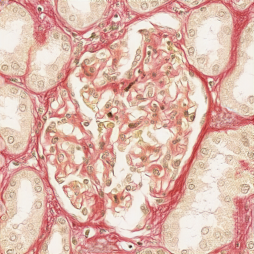}&
			\includegraphics[width=0.14\textwidth]{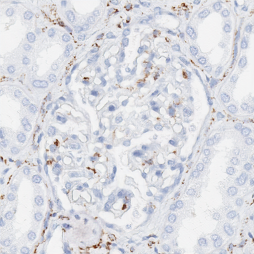}&
			\includegraphics[width=0.14\textwidth]{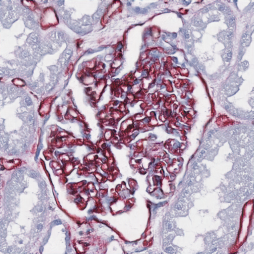}&
			
			\\
			& \rowname{Seg.}&
			\includegraphics[width=0.14\textwidth]{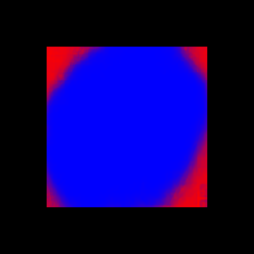}&
			\includegraphics[width=0.14\textwidth]{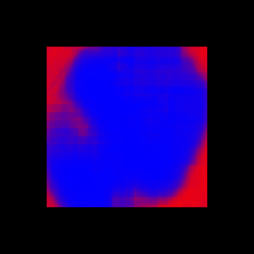}&
			\includegraphics[width=0.14\textwidth]{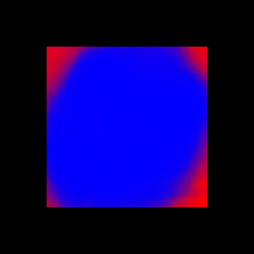}&
			\includegraphics[width=0.14\textwidth]{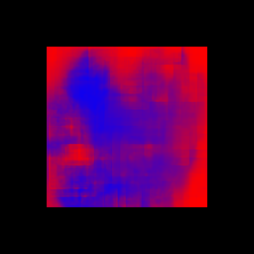}&
			\includegraphics[width=0.14\textwidth]{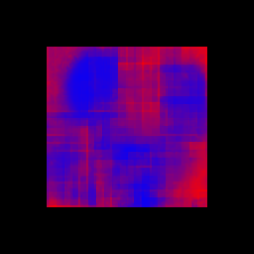}&
			\includegraphics[width=0.14\textwidth]{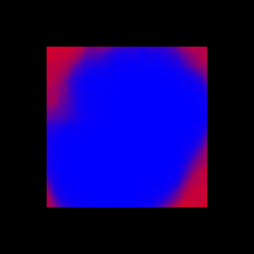}&
			
			\\	
			\multirow{2}{*}{\rowname{HistoStarGAN}} & \rowname{}&		
			\includegraphics[width=0.14\textwidth]{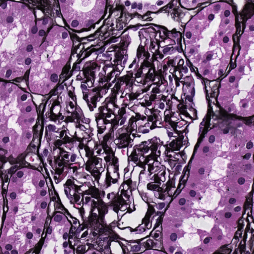}&
			\includegraphics[width=0.14\textwidth]{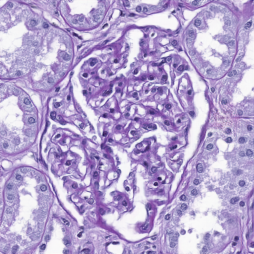}&
			\includegraphics[width=0.14\textwidth]{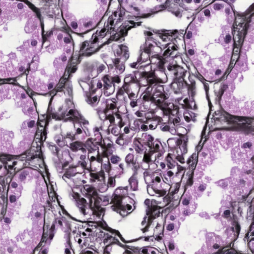}&
			\includegraphics[width=0.14\textwidth]{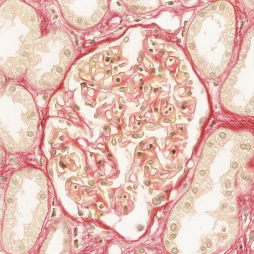}&
			\includegraphics[width=0.14\textwidth]{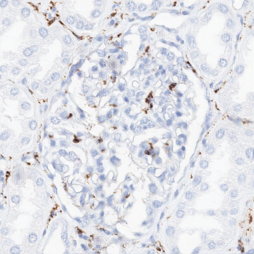}&
			\includegraphics[width=0.14\textwidth]{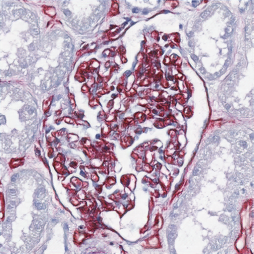}&
			
			\\
			& \rowname{Seg.}&
			\includegraphics[width=0.14\textwidth]{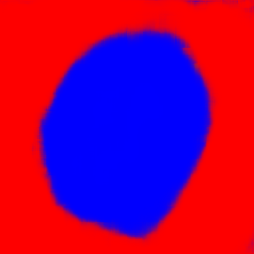}&
			\includegraphics[width=0.14\textwidth]{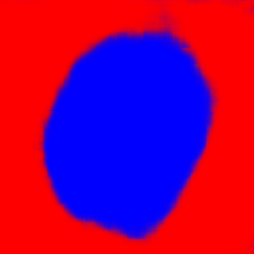}&
			\includegraphics[width=0.14\textwidth]{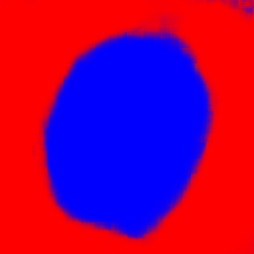}&
			\includegraphics[width=0.14\textwidth]{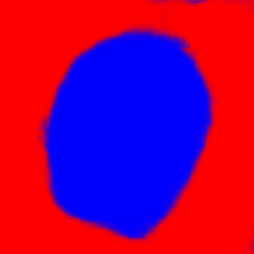}&
			\includegraphics[width=0.14\textwidth]{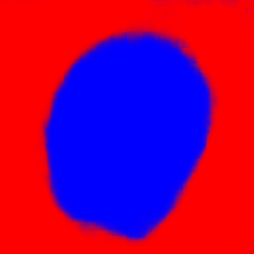}&
			\includegraphics[width=0.14\textwidth]{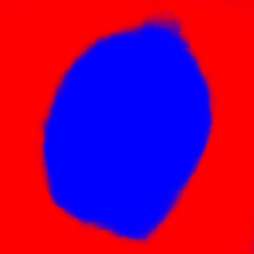}&

			\\
			\multirow{2}{*}{\rowname{StarGANv2}} & \rowname{}&			
			\includegraphics[width=0.14\textwidth]{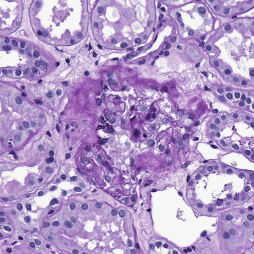}&
			\includegraphics[width=0.14\textwidth]{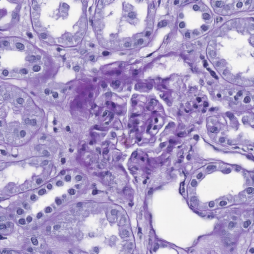}&
			\includegraphics[width=0.14\textwidth]{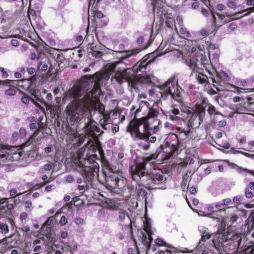}&
			\includegraphics[width=0.14\textwidth]{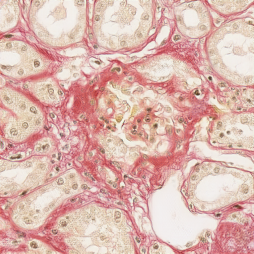}&
			\includegraphics[width=0.14\textwidth]{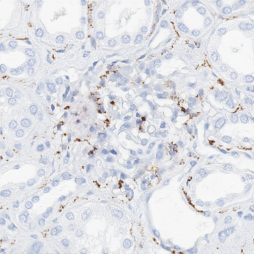}&
			\includegraphics[width=0.14\textwidth]{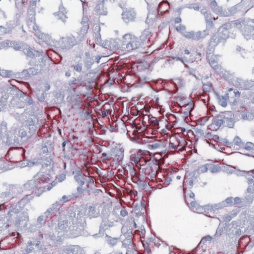}&
			\\
			& \rowname{Seg.}		&	
			\includegraphics[width=0.14\textwidth]{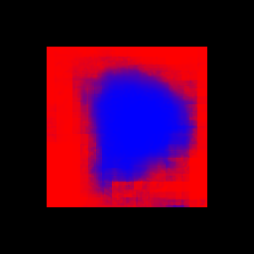}&
			\includegraphics[width=0.14\textwidth]{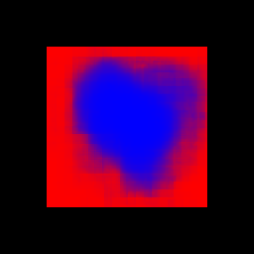}&
			\includegraphics[width=0.14\textwidth]{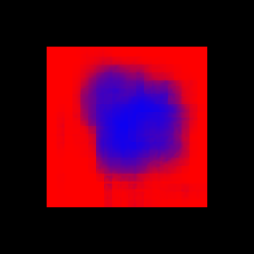}&
			\includegraphics[width=0.14\textwidth]{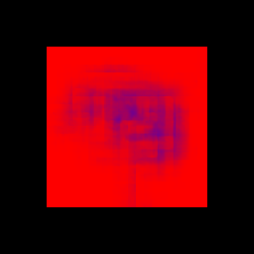}&
			\includegraphics[width=0.14\textwidth]{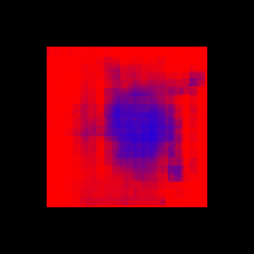}&
			\includegraphics[width=0.14\textwidth]{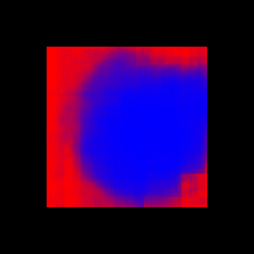}&

			\\
			\multirow{2}{*}{\rowname{HistoStarGAN}} & \rowname{}&					
			\includegraphics[width=0.14\textwidth]{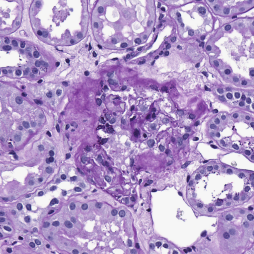}&
			\includegraphics[width=0.14\textwidth]{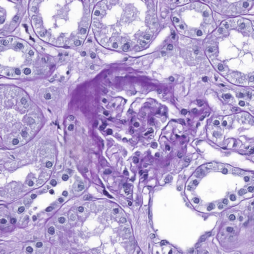}&
			\includegraphics[width=0.14\textwidth]{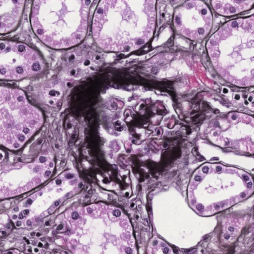}&
			\includegraphics[width=0.14\textwidth]{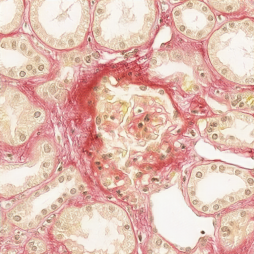}&
			\includegraphics[width=0.14\textwidth]{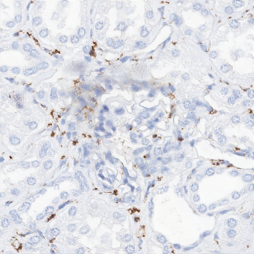}&
			\includegraphics[width=0.14\textwidth]{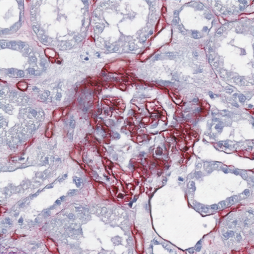}&
			\\
			& \rowname{Seg.}	&					
			\includegraphics[width=0.14\textwidth]{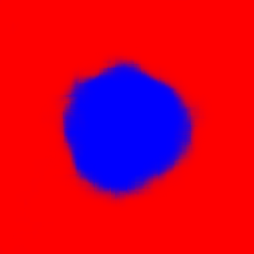}&
			\includegraphics[width=0.14\textwidth]{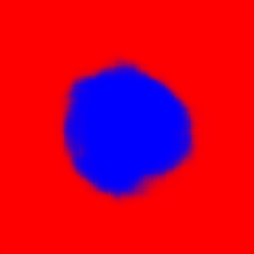}&
			\includegraphics[width=0.14\textwidth]{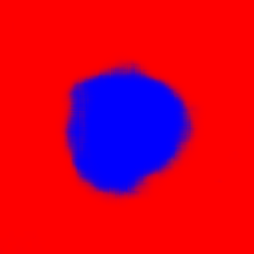}&
			\includegraphics[width=0.14\textwidth]{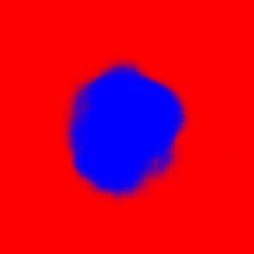}&
			\includegraphics[width=0.14\textwidth]{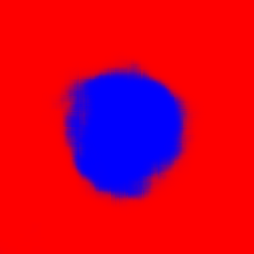}&
			\includegraphics[width=0.14\textwidth]{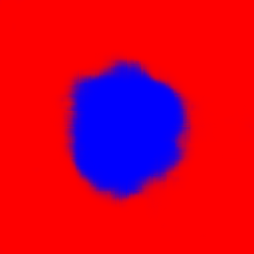}&
			
		\end{tabular}
		\caption[Visual comparison between StarGANv2 and HistoStarGAN models trained on balanced dataset]{StarGANv2 trained on a balanced CycleGAN-generated dataset compared to  HistoStarGAN.  The segmentations for StarGANv2 are obtained by stain-specific, pre-trained baseline models. Each translation and segmentation is averaged over 50 random latent codes. N.B: HistoStarGAN yields more accurate segmentations. Pre-trained models result in $324 \times 324$ pixel images which are placed in the centre of a $512 \times 512$ pixel black squares.}
		\label{fig:histostarganv2_vs_starganv2_seg}
	\end{center}
\end{figure}

\paragraph{CycleGAN-Generated Balanced Dataset}

A StarGANv2 model trained on  a  balanced dataset produced by CycleGAN translations of the annotated stain,  can also result in diverse multi-domain translations between multiple stainings, as illustrated in Figure \ref{fig:starganv2_cyclegan_dataset_other_translation}. Usually, the glomeruli structures are visually preserved. However, when measuring the quality of obtained translation  using the  performances of pre-trained segmentation models for each staining (baseline models trained in a fully-supervised setting), the conclusion can be different. Figure \ref{fig:histostarganv2_vs_starganv2_seg} presents a  visual comparisons of HistoStarGAN and StarGANv2, where translations obtained by StarGANv2 are segmented using pre-trained models from each staining. Since both models can obtain diverse translations, Figure \ref{fig:histostarganv2_vs_starganv2_seg} represent the average translation and average segmentation over 50 random latent codes. It can be seen that HistoStarGAN  gives a  more accurate segmentations, especially in difficult cases, e.g.\ sclerotic glomeruli, rows 5 and 7. Nevertheless, in the absence of a segmentation branch, the dataset composition itself cannot be a strong guarantee that glomeruli structures will be preserved. Thus, an explicit requirement, such as attaching the proposed segmentation branch, is beneficial to ensure the correctness of the translation and robust segmentation. 

\begin{figure}[!htbp]
	\begin{center}
		\settoheight{\tempdima}{\includegraphics[width=0.145\textwidth]{Images_for_HistoStarGANv2/CycleGAN_whole_dataset/03_original.png}}%
		\begin{tabular}{@{}c@{ }c@{ }c@{ }c@{ }c@{ }c@{ }c@{ }c@{ }c@{ }c@{ }}
			& & \multicolumn{5}{c}{Translations}\\
			\cline{3-7}\vspace{-1.5ex}\\
			& Real &PAS &Jones H\&E & Sirius Red & CD68 & CD34\\
			\rowname{PAS} &
			\includegraphics[width=0.145\textwidth]{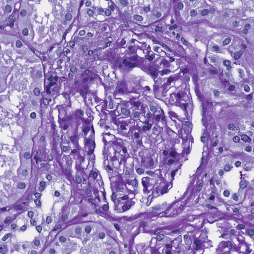}&
			\includegraphics[width=0.145\textwidth]{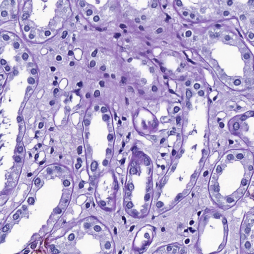}&
			\includegraphics[width=0.145\textwidth]{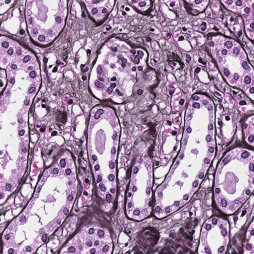} &
			\includegraphics[width=0.145\textwidth]{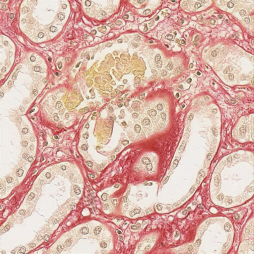} &
			\includegraphics[width=0.145\textwidth]{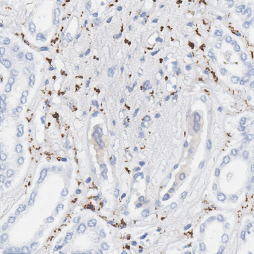} &
			\includegraphics[width=0.145\textwidth]{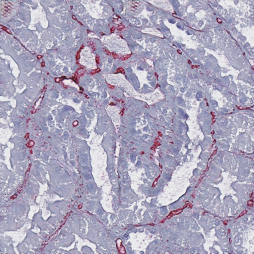}
			\\
			\rowname{Jones H\&E} &
			\includegraphics[width=0.145\textwidth]{Images_for_HistoStarGANv2/CycleGAN_whole_dataset/03_original.png}&
			\includegraphics[width=0.145\textwidth]{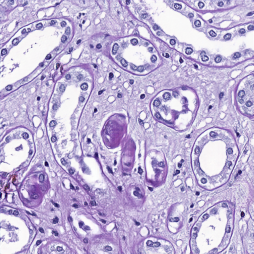}&
			\includegraphics[width=0.145\textwidth]{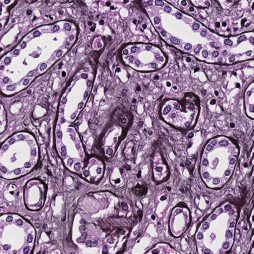} &
			\includegraphics[width=0.145\textwidth]{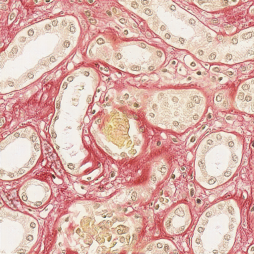} &
			\includegraphics[width=0.145\textwidth]{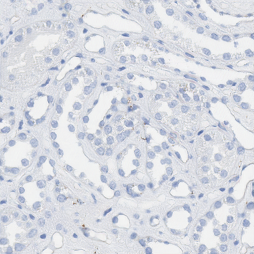} &
			\includegraphics[width=0.145\textwidth]{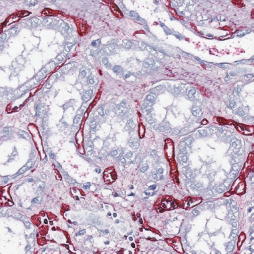}
			\\
			\rowname{CD68} &
			\includegraphics[width=0.145\textwidth]{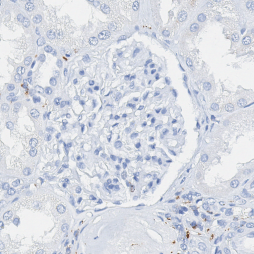}&
			\includegraphics[width=0.145\textwidth]{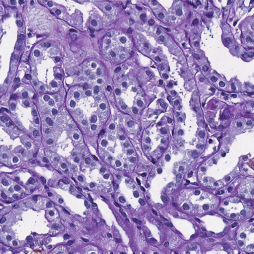}&
			\includegraphics[width=0.145\textwidth]{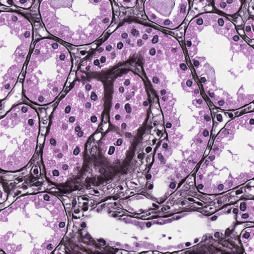} &
			\includegraphics[width=0.145\textwidth]{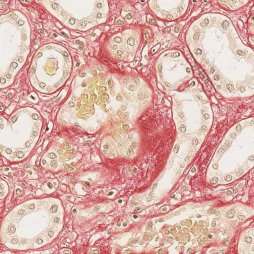} &
			\includegraphics[width=0.145\textwidth]{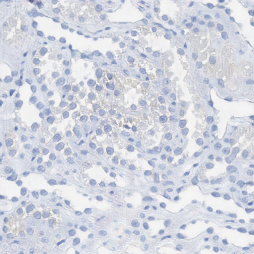} &
			\includegraphics[width=0.145\textwidth]{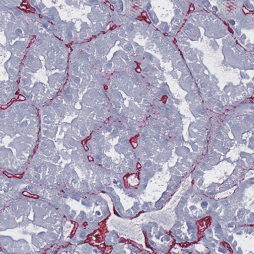}
			\\
			\rowname{Sirius Red} &
			\includegraphics[width=0.145\textwidth]{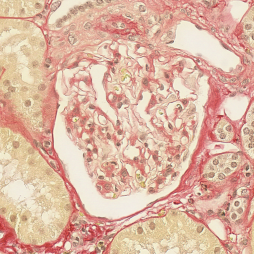}&
			\includegraphics[width=0.145\textwidth]{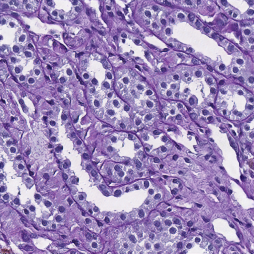}&
			\includegraphics[width=0.145\textwidth]{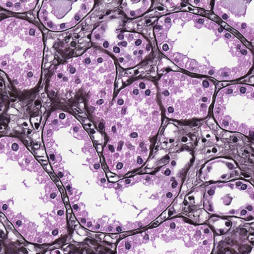} &
			\includegraphics[width=0.145\textwidth]{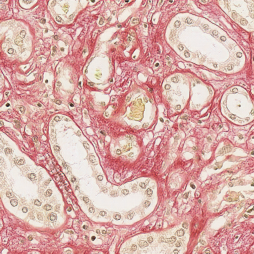} &
			\includegraphics[width=0.145\textwidth]{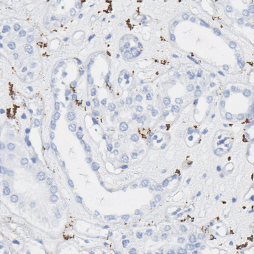} &
			\includegraphics[width=0.145\textwidth]{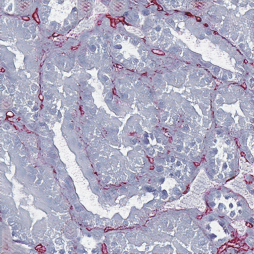}
			\\
			\rowname{CD34} &
			\includegraphics[width=0.145\textwidth]{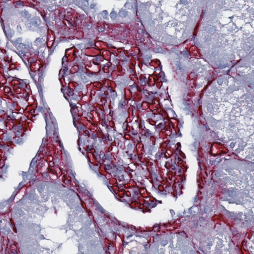}&
			\includegraphics[width=0.145\textwidth]{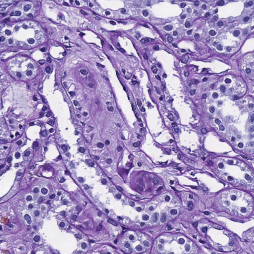}&
			\includegraphics[width=0.145\textwidth]{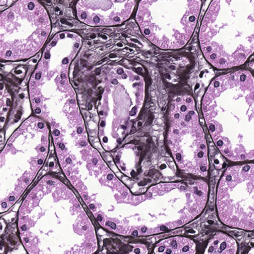} &
			\includegraphics[width=0.145\textwidth]{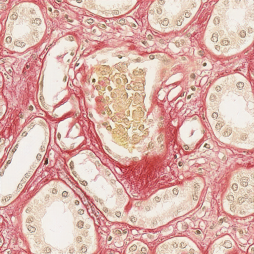} &
			\includegraphics[width=0.145\textwidth]{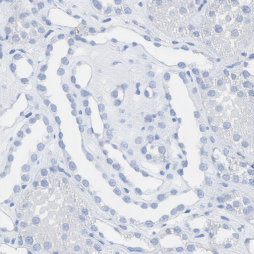} &
			\includegraphics[width=0.145\textwidth]{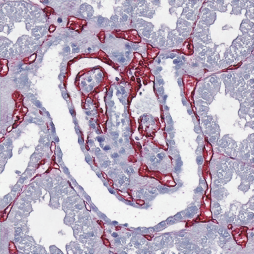}
			\\			
		\end{tabular}
		\caption[StarGANv2 model trained on imbalanced dataset]{StarGANv2 model trained on CycleGAN-generated translations of an imbalanced  PAS  dataset to other stains. Each image is generated using different latent codes.}
		\label{fig:starganv2_cyclegan_whole_dataset_other_translation}
	\end{center}
\end{figure}

\paragraph{CycleGAN-Generated Imbalanced Dataset}
If the dataset used to train the StarGANv2 is imbalanced, the model can no longer preserve the structure of interest. Some examples of stain transfers obtained with this model are presented in Figure  \ref{fig:starganv2_cyclegan_whole_dataset_other_translation}. Since the HistoStarGAN model trained using the same imbalanced dataset does not have such a problem, this demonstrates the significance of the proposed segmentation branch.

\begin{figure}[!htbp]
	\begin{center}
		\settoheight{\tempdima}{\includegraphics[width=0.145\textwidth]{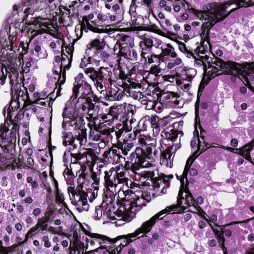}}%
		\begin{tabular}{@{}c@{ }c@{ }c@{ }c@{ }c@{ }c@{ }c@{ }c@{ }c@{ }c@{ }}
			& & \multicolumn{5}{c}{Translations}\\
			\cline{3-7}\vspace{-1.5ex} \\
			& Real &PAS &Jones H\&E & Sirius Red & CD68 & CD34\\ 
			\rowname{PAS} &
			\includegraphics[width=0.145\textwidth]{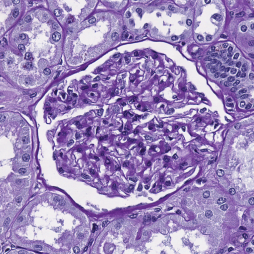}&
			\includegraphics[width=0.145\textwidth]{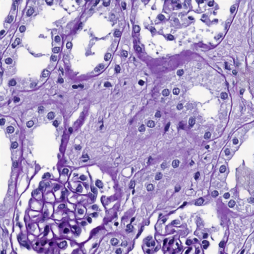} &
			\includegraphics[width=0.145\textwidth]{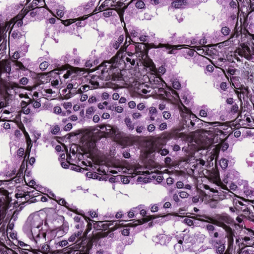} &
			\includegraphics[width=0.145\textwidth]{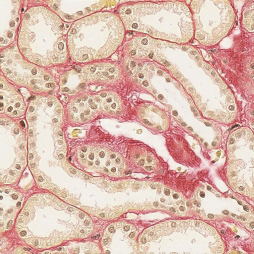} &
			\includegraphics[width=0.145\textwidth]{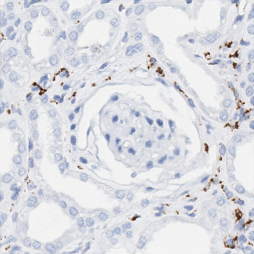} &
			\includegraphics[width=0.145\textwidth]{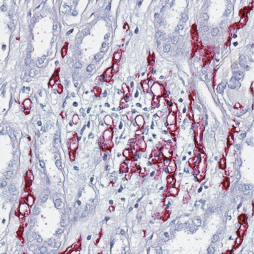} \\
			\rowname{Jones H\&E} &
			\includegraphics[width=0.145\textwidth]{Images_for_HistoStarGANv2/Baseline/others_to_others/03_original.png}&
			\includegraphics[width=0.145\textwidth]{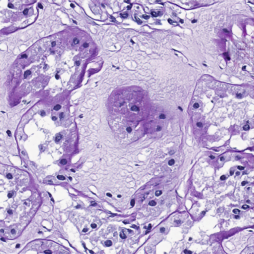}&
			\includegraphics[width=0.145\textwidth]{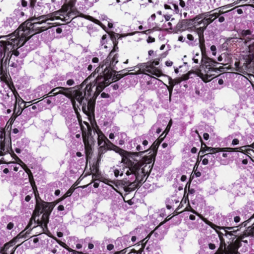} &
			\includegraphics[width=0.145\textwidth]{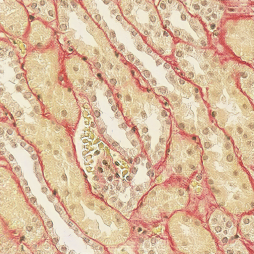} &
			\includegraphics[width=0.145\textwidth]{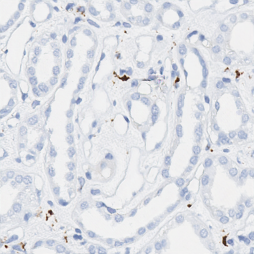} &
			\includegraphics[width=0.145\textwidth]{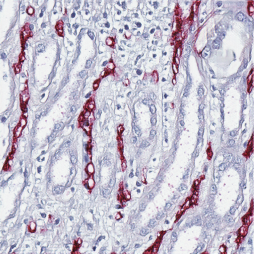}
			\\
			\rowname{CD68} &
			\includegraphics[width=0.145\textwidth]{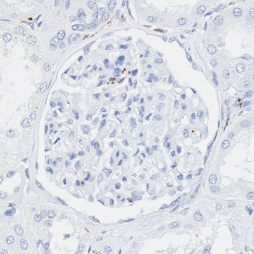}&
			\includegraphics[width=0.145\textwidth]{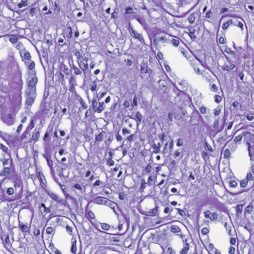}&
			\includegraphics[width=0.145\textwidth]{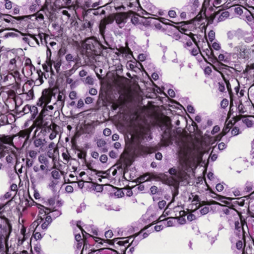} &
			\includegraphics[width=0.145\textwidth]{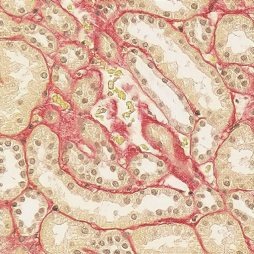} &
			\includegraphics[width=0.145\textwidth]{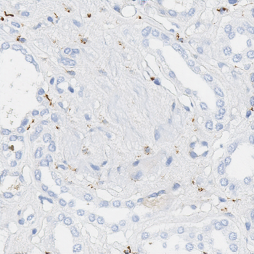} &
			\includegraphics[width=0.145\textwidth]{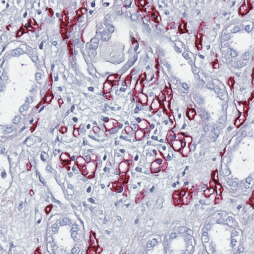}
			\\
			\rowname{Sirius Red} &
			\includegraphics[width=0.145\textwidth]{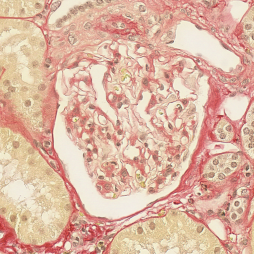}&
			\includegraphics[width=0.145\textwidth]{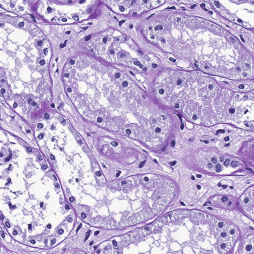}&
			\includegraphics[width=0.145\textwidth]{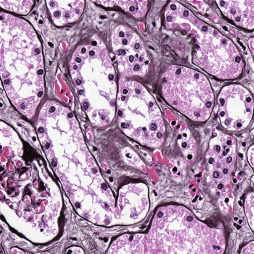} &
			\includegraphics[width=0.145\textwidth]{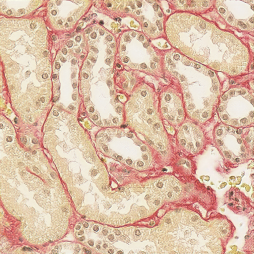} &
			\includegraphics[width=0.145\textwidth]{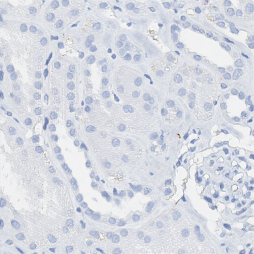} &
			\includegraphics[width=0.145\textwidth]{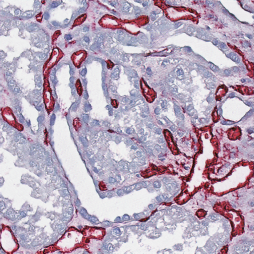}
			\\
			\rowname{CD34} &
			\includegraphics[width=0.145\textwidth]{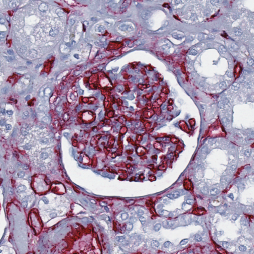}&
			\includegraphics[width=0.145\textwidth]{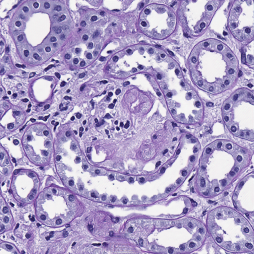}&
			\includegraphics[width=0.145\textwidth]{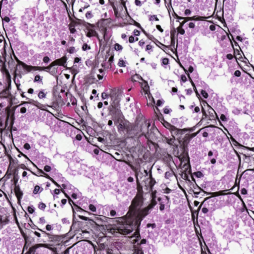} &
			\includegraphics[width=0.145\textwidth]{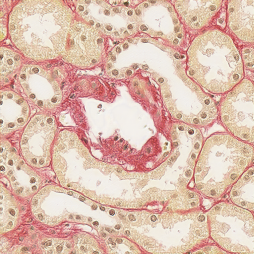} &
			\includegraphics[width=0.145\textwidth]{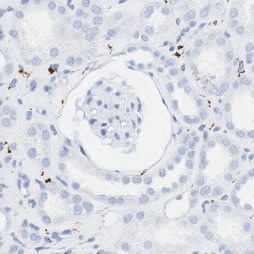} &
			\includegraphics[width=0.145\textwidth]{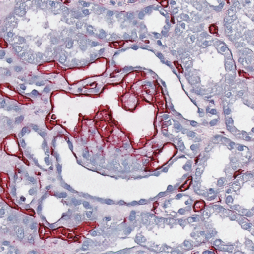}
			\\			
		\end{tabular}
		\caption[StarGANv2 model trained on randomly extracted dataset]{StarGANv2 model trained on random patches extracted from all stains, applied to  translated glomeruli patches. The images in each column are generated using different latent codes.}
		\label{fig:starganv2_random_original_translation}
	\end{center}
\end{figure}

\paragraph{Random Dataset:} StarGANv2 can also be trained on a dataset composed of random patches extracted from each stain (PAS, Jones H\&E, Sirius Red, CD68 and CD34). Thus, \num{5000} random patches from the training patients are extracted uniformly, and the total dataset containing \num{25000} images is used to train the StarGANv2 model. As in the CycleGAN-generated imbalanced data setting, this model cannot preserve the structure of interest, as demonstrated in Figure \ref{fig:starganv2_random_original_translation}. Moreover, it is prone to bigger alterations of microscopic structures, which limits the usefulness of such translations and confirms the benefits of the proposed HistoStarGAN model.

\section{KidneyArtPathology Dataset}
\label{sec:chapter_histostargan:sec_kidneyArtPathology_dataset}
\begin{figure}[!htbp]
	\begin{center}
		\settoheight{\tempdima}{\includegraphics[width=0.15\textwidth]{Images_for_HistoStarGANv2/CycleGAN/02_orig.png}}%
		\begin{tabular}{@{}c@{ }c@{ }c@{ }c@{ }c@{ }c@{ }c@{ }c@{ }c@{ }c@{ }}
			\multicolumn{5}{c}{Generated images}\\
			\cline{1-5 \vspace{-1.5ex}}\\
			PAS &Jones H\&E & Sirius Red & CD68 & CD34\\
			\includegraphics[width=0.15\textwidth]{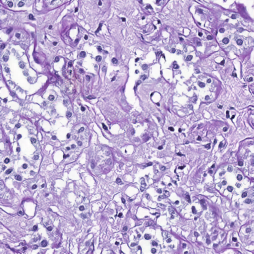}&
			\includegraphics[width=0.15\textwidth]{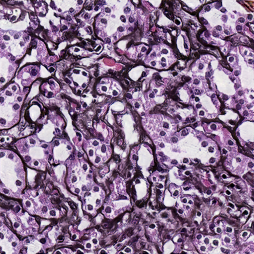}&			\includegraphics[width=0.15\textwidth]{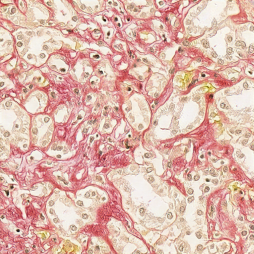}&
			\includegraphics[width=0.15\textwidth]{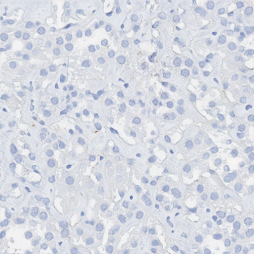}&
			\includegraphics[width=0.15\textwidth]{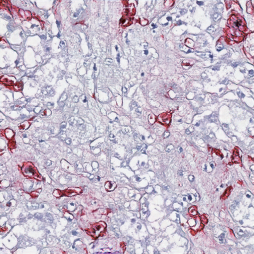}&
			
			\\
			\includegraphics[width=0.15\textwidth]{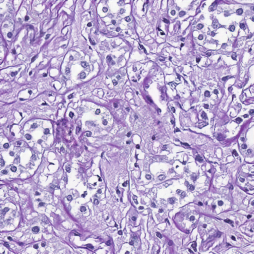}&
			\includegraphics[width=0.15\textwidth]{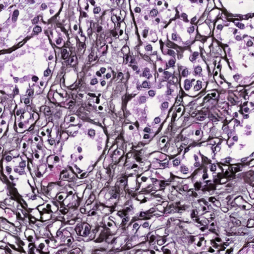}&
			\includegraphics[width=0.15\textwidth]{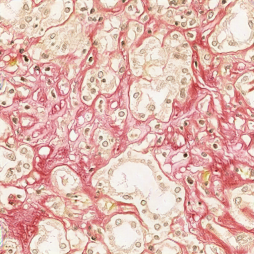}&
			\includegraphics[width=0.15\textwidth]{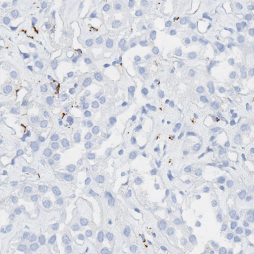}&
			\includegraphics[width=0.15\textwidth]{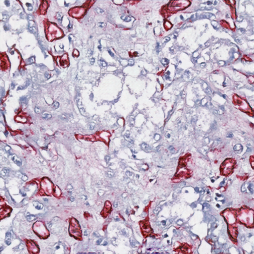}&
			\\
			\includegraphics[width=0.15\textwidth]{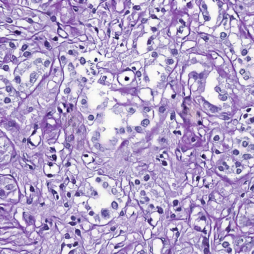}&
			\includegraphics[width=0.15\textwidth]{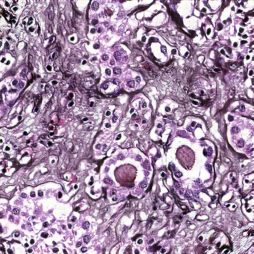}&		\includegraphics[width=0.15\textwidth]{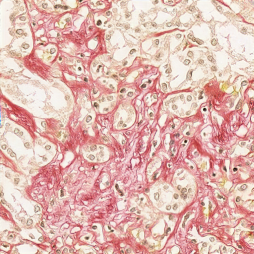}&
			\includegraphics[width=0.15\textwidth]{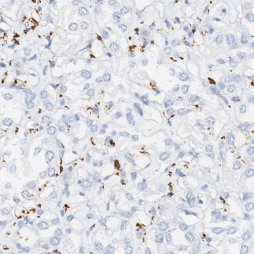}&
			\includegraphics[width=0.15\textwidth]{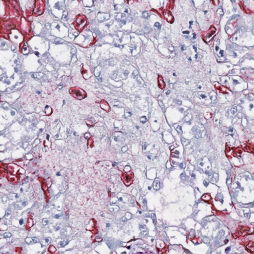}&
			
		\end{tabular}
		\caption[KidneyArtPathology - Images generated from random noise]{The HistoStarGAN model generates plausible histopathological images from random noise.}
		\label{fig:histostargan_new_histopathological_images}
	\end{center}
\end{figure}
\begin{figure}[!htbp]
	\begin{center}
		\settoheight{\tempdima}{\includegraphics[width=0.15\textwidth]{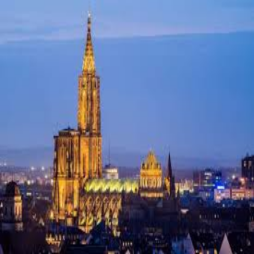}}%
		\begin{tabular}{@{}c@{ }c@{ }c@{ }c@{ }c@{ }c@{ }c@{ }c@{ }c@{ }}
			& \multicolumn{5}{c}{Translations}\\
			\cline{2-7} \vspace{-1.5ex}\\
			Image &PAS &Jones H\&E & Sirius Red & CD68 & CD34\\
			\includegraphics[width=0.15\textwidth]{Images_for_HistoStarGANv2/HistoStarGAN_balanced_dataset_w_5/natural_images/natural_images_strasbourg_original.png}&
			\includegraphics[width=0.15\textwidth]{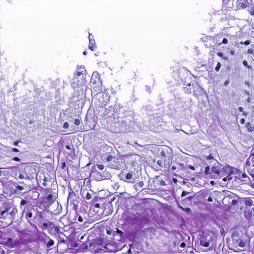}&
			\includegraphics[width=0.15\textwidth]{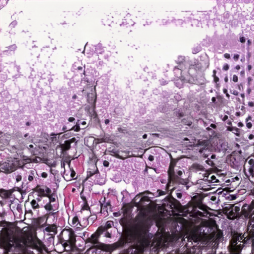}&
			\includegraphics[width=0.15\textwidth]{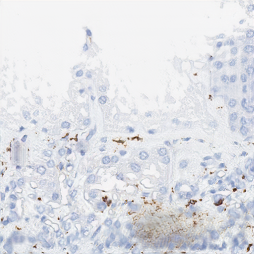}&
			\includegraphics[width=0.15\textwidth]{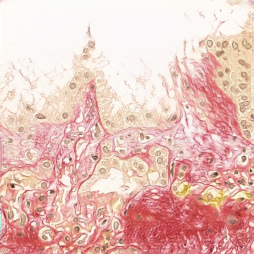}&
			\includegraphics[width=0.15\textwidth]{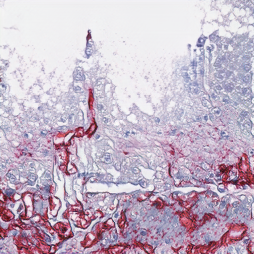}&
			\\
			\includegraphics[width=0.15\textwidth]{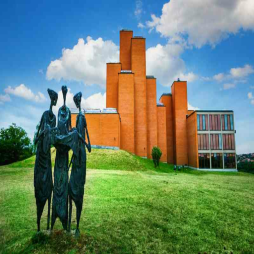}&
			\includegraphics[width=0.15\textwidth]{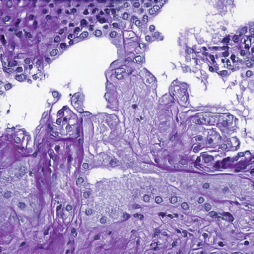}&
			\includegraphics[width=0.15\textwidth]{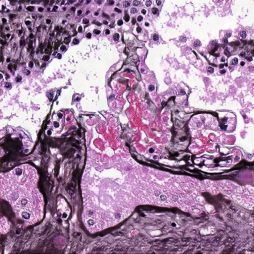}&
			\includegraphics[width=0.15\textwidth]{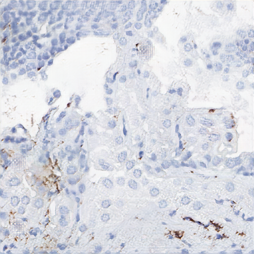}&
			\includegraphics[width=0.15\textwidth]{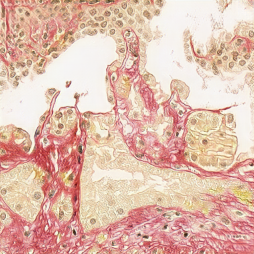}&
			\includegraphics[width=0.15\textwidth]{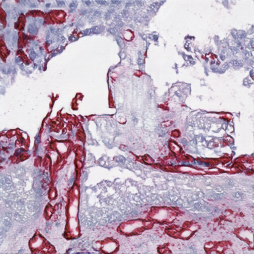}&
			
			\\
			\includegraphics[width=0.15\textwidth]{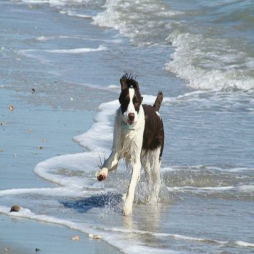}&
			\includegraphics[width=0.15\textwidth]{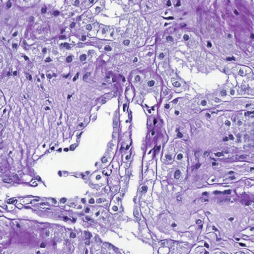}&
			\includegraphics[width=0.15\textwidth]{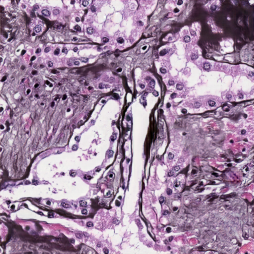}&
			\includegraphics[width=0.15\textwidth]{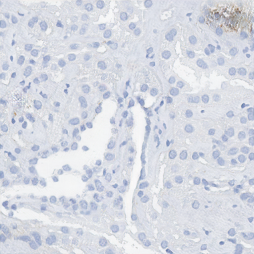}&
			\includegraphics[width=0.15\textwidth]{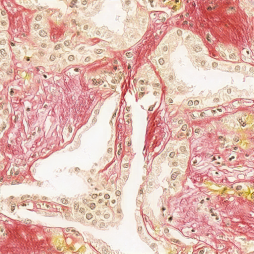}&
			\includegraphics[width=0.15\textwidth]{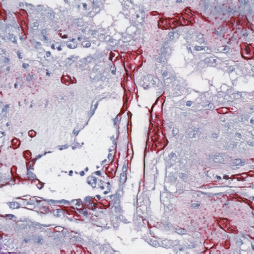}&
		\end{tabular}
		\caption[KidneyArtPathology - Images generated from natural images]{KidneyArtPathology - histopathological image generation from natural images\protect\footnotemark.}
		\label{fig:kidney_art_pathology_natural_images}
	\end{center}
\end{figure}
\footnotetext{Image credits in order of appearance: Strasbourg Cathedral:  https://www.visitstrasbourg.fr/en/discover/must-see-attractions/the-cathedral/, Kragujevac Museum:  https://www.topworldtraveling.com/articles/travel-guides/15-best-things-to-do-in-kragujevac-serbia.html, Dog: https://github.com/fastai/imagenette}


\color{black} Here we describe a new dataset released to encourage the progress of deep learning-based solutions in the field of renal pathology. The dataset contains 5000 images from five stainings, in a resolution of $512\times512$ pixels, fully annotated for the task of glomeruli segmentation. To achieve this, a HistoStarGAN model is  trained using a dataset that is as representative as possible, i.e.\ two immunohistochemical stainings (CD68 and CD34) in addition to three histochemical stainings (PAS, Jones H\&E and Sirius Red). Thus, the model can generate various translations in multiple stainings,  which has allowed the creation of the KidneyArtPathology dataset \footnote{KidneyArtPathology has been release online, {\href{https://main.d33ezaxrmu3m4a.amplifyapp.com/}{ https://main.d33ezaxrmu3m4a.amplifyapp.com/}}}. The dataset is composed of HistoStarGAN translations of PAS stain images (the same as those used to generate the annotated dataset used for HistoStarGAN training) which are translated using 10 random latent vectors into each stainings, including PAS.


Moreover, the associated pre-trained HistoStarGAN model (also publicly available) can be further used to augment private datasets by augmenting the number of stainings to those used during HistStarGAN's training.
There are several benefits of such a dataset, which fall under three categories:
\begin{itemize}
	\item Pathological -- Non-invasive Pathology Training, the diverse appearance of glomeruli can be helpful in the early stages of a pathologist's training. 
	\item Benchmarks -- the absence of publicly available real-world datasets poses huge challenges for  rigorous comparison in the literature. Thus, such a large collection of annotated patches can serve as a benchmark.
	\item Domain adaptation -- The data can be used in addition to a private dataset (which can contain limited data) to build more robust models, e.g.\ as an augmentation or domain adaptation strategy.
\end{itemize}

\textbf{New histopathological images: }Since the style of a stain is encoded by the mapping network, it is possible to generate new patches in each training staining by providing a random image, rather than a source histopathological patch,  to be translated to a given stain. In addition to a fully-annotated glomeruli dataset, new histopathological images in different stains can be generated. Some generated examples are provided in Figure \ref{fig:histostargan_new_histopathological_images}. Alternatively, by providing  non-histopathological image, the HistoStarGAN is able to convert it to a histopathological image, examples of which are provided in Figure \ref{fig:kidney_art_pathology_natural_images} (although medical use of this is admittedly most likely limited to non-existent, it is an interesting side property of HistoStarGAN) . 

\section{Limitations and Opportunities}
\label{sec:limitations_and_opportunities}
\begin{figure}[!htbp]
	\begin{center}
		\begin{tabular}{c@{ }c@{ }}
			\begin{tikzpicture}[zoomboxarray,zoomboxarray columns=1,zoomboxarray rows=1]
				\node [image node] { \includegraphics[width=0.25\textwidth]{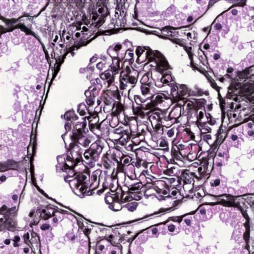} };
				\zoombox[color code=red,magnification=3]{0.43,0.61}
			\end{tikzpicture}	\vspace{-1.5em} \\
				\begin{tikzpicture}[zoomboxarray,zoomboxarray columns=1,zoomboxarray rows=1]
				\node [image node] { \includegraphics[width=0.25\textwidth]{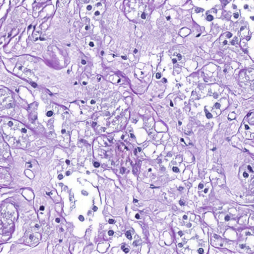} };
				\zoombox[color code=red,magnification=3]{0.49,0.4}
			\end{tikzpicture}	\vspace{0em} \\
		\end{tabular}
		\caption[HistoStarGAN - common artefacts]{HistoStarGAN - common artefacts. N.B. the artefacts are well-incorporated into the image texture and not obviously noticeable at first glance.}
		\label{fig:histostargan_failure_case}
	\end{center}
\end{figure}

HistoStarGAN is an end-to-end model that can segment and generate diverse plausible histopathological images alongside their segmentation masks. However, for some latent codes the model can produce specific artefacts in the translations. These are usually well-incorporated into the overall texture of the image, which makes them not obvious at first glance. A closer look at one such example is given in  Figure \ref{fig:histostargan_failure_case}. The primary hypothesis is that the discriminator does not have enough capacity to spot these artefacts, and thus a more sophisticated discriminator could  be considered.

Furthermore, the performance of the HistoStarGAN model can be affected by the choice of source and target stainings and the quality of CycleGAN translations, which remains to be explored in future work.  HistoStarGAN is based on CycleGAN translations, which are proven to be noisy \cite{vasiljevic_isbi_2021,vasiljevicCYCLEGANVIRTUALSTAIN2021}, and therefore some of these invisible artefacts may be recreated by HistoStarGAN. Thus, it is unclear what information from the source domain is preserved in translations and whether any of the applied augmentations can perturb them. It is possible that incorporating additional real examples from the target stainings can improve translation quality.
Additionally, it has been confirmed experimentally that the choice of loss functions used for the diversity loss can significantly affect the HistoStarGAN translations. For example, the model will be forced to perform structural changes if diversity loss is set to maximise the structural similarity between images (this is illustrated on the project's website linked to above). Although from a style-transfer perspective, such a solution is interesting, changing internal structures in this way is not biologically justifiable. However, according to a recent study \cite{yamashita2021learning}, medically-irrelevant augmentation can increase model robustness; thus, this represents another exciting direction for the work ahead. 

Nevertheless, the model has great potential to be used for tackling the lack of annotations in histopathological datasets. Depending on the availability of labels in a given target domain, diverse HistoStarGAN translations with corresponding segmentation masks can help training from unsupervised to fully supervised settings, using one or more source (annotated) domains. 

Finally, it is crucial to note that translations obtained by the HistoStarGAN model should not be used for diagnostic purposes. All images are artificially generated, and the model can perturb diagnostic markers. Thus, the dataset composed of HistoStarGAN translations should only be used for general-purpose analysis related to glomeruli (e.g.\ counting).

\section{Conclusions}
\label{sec:conclusion}
The HistoStarGAN model represents the first end-to-end trainable solution for simultaneous stain normalisation, stain transfer and stain invariant segmentation. For the first time, obtaining highly plausible stain transfer from unseen stainings is possible without any additional change to the model (e.g.\ fine-tuning). Moreover, the model achieves new state-of-the-art results in stain invariant segmentation, successfully generalising to six unseen stainings. The proposed solution is general and extendible to new stainings or use cases.

Being able to generate diverse translations for a given input, the proposed solution paved the way to generate KidneyArtPathology, an artificially created and fully annotated data set illustrating a broad spectrum of synthetic image data ranging from biologically usable results, to visually plausible but biologically useless output. 

However, the obtained virtual staining can realistically change the appearance of microscopic structures, such as the number of visible nuclei, size of Bowman's capsule or membrane thickness. Thus, it is ill-advised to use these translations for applications that rely on the assumption that the translation process preserves all structures in the original image. Moreover, such virtually stained images must not be used for diagnostic purposes since diagnostic markers can be altered. As such, this study goes towards saving resources and time for annotating the high-quality datasets required for developing deep learning based models, not to replace the process of physical staining.

Nevertheless, it remains an open question why the model's performance varies across different stainings. One hypothesis is that the model, being based on CycleGAN translations, replicates some of their limitations. However, the influence of these  CycleGAN models and the choice of the stainings used in the dataset remains to be explored in future work.

\section*{Acknowledgements}
This work was supported by: ERACoSysMed and e:Med initiatives by the German Ministry of Research and Education (BMBF); SysMIFTA (project management PTJ, FKZ 031L-0085A; Agence National de la Recherche, ANR, project number ANR-15—CMED-0004); SYSIMIT (project management DLR, FKZ 01ZX1608A); ArtIC project ''Artificial Intelligence for Care'' (grant ANR-20-THIA-0006-01) and co funded by Région Grand Est, Inria Nancy - Grand Est, IHU of Strasbourg, University of Strasbourg and University of Haute-Alsace; and the French Government through co-tutelle PhD funding.
We thank the Nvidia Corporation, the \emph{Centre de Calcul de l'Université de Strasbourg}, and HPC resources of IDRIS under the allocation 2020-A0091011872 made by GENCI. We also thank the MHH team for providing high-quality images and annotations, specifically Nicole Kroenke for excellent technical assistance, Nadine Schaadt for image management and quality control, and Valery Volk and Jessica Schmitz for annotations under the supervision of domain experts.

\bibliography{main}
\end{document}